\documentclass[journal]{IEEEtran}
\ifCLASSINFOpdf
   \usepackage[pdftex]{graphicx}
   \graphicspath{{../pdf/}{../jpeg/}}
   \DeclareGraphicsExtensions{.pdf,.jpeg,.png}
\else
   \usepackage[dvips]{graphicx}
   \graphicspath{{../eps/}}
   \DeclareGraphicsExtensions{.eps}
\fi
\usepackage[cmex10]{amsmath}
\usepackage{color}
\usepackage{algorithmic}
\usepackage{algorithm}
\usepackage{multirow}
\usepackage[misc]{ifsym}
\begin{document}
\title{BIT: Biologically Inspired Tracker}

\author{Bolun~Cai,
        Xiangmin~Xu,~\IEEEmembership{Member,~IEEE,}
        Xiaofen~Xing,~\IEEEmembership{Member,~IEEE,}
        Kui~Jia,~\IEEEmembership{Member,~IEEE,}
        Jie~Miao,
        and~Dacheng~Tao,~\IEEEmembership{Fellow,~IEEE}
\thanks{B. Cai, X. Xu (\Letter), X. Xing and J. Miao are with the School of Electronic and Information Engineering, South China University of Technology, Guangzhou, Guangdong, China
 (e-mail: caibolun@gmail.com; xmxu@scut.edu.cn; xfxing@scut.edu.cn; miaow1988@qq.com).}
 \thanks{K. Jia is with the department of Computer and Information Science, Faculty of Science and Technology, University of Macau, E11 Avenida da Universidade, Taipa, Macau SAR, China.
 (e-mail: kuijia@gmail.com).}
\thanks{D. Tao is with the Centre for Quantum Computation \& Intelligent Systems and the Faculty of Engineering and Information Technology, University of Technology Sydney, 81 Broadway Street, Ultimo, NSW 2007, Australia (e-mail: dacheng.tao@uts.edu.au).}\thanks{\textcircled{c} 2019 IEEE. Personal use of this material is permitted. Permission from IEEE must be obtained for all other uses, in any current or future media, including reprinting/republishing this material for advertising or promotional purposes, creating new collective works, for resale or redistribution to servers or lists, or reuse of any copyrighted component of this work in other works.}}

\maketitle

\begin{abstract}
Visual tracking is challenging due to image variations caused by various factors, such as object deformation, scale change, illumination change and occlusion. Given the superior tracking performance of human visual system (HVS), an ideal design of biologically inspired model is expected to improve computer visual tracking. This is however a difficult task due to the incomplete understanding of neurons' working mechanism in HVS. This paper aims to address this challenge based on the analysis of visual cognitive mechanism of the ventral stream in the visual cortex, which simulates shallow neurons (S1 units and C1 units) to extract low-level biologically inspired features for the target appearance and imitates an advanced learning mechanism (S2 units and C2 units) to combine generative and discriminative models for target location. In addition, fast Gabor approximation (FGA) and fast Fourier transform (FFT) are adopted for real-time learning and detection in this framework. Extensive experiments on large-scale benchmark datasets show that the proposed biologically inspired tracker performs favorably against state-of-the-art methods in terms of efficiency, accuracy, and robustness. The acceleration technique in particular ensures that BIT maintains a speed of approximately 45 frames per second.
\end{abstract}


\IEEEpeerreviewmaketitle

\section{Introduction}\label{sec:Introduction}

\IEEEPARstart{V}{isual} object tracking is a fundamental problem in computer vision. It has a wide variety of applications \cite{review} including motion analysis, video surveillance, human computer interaction and robot perception. Although visual tracking has been intensively investigated in the past decade, it is still challenging caused by various factors such as appearance variations, pose change, occlusion. To improve visual tracking, one may need to address all of these challenges by developing better feature representations of visual targets and more effective tracking models.

Target objects in visual tracking are commonly represented as handcrafted features or automated features. Histogram based handcrafed features have been introduced to tracking, such as a color histogram \cite{kms} embedded in the mean-shift algorithm to search the target, a histogram of oriented gradients (HOG) \cite{kcf} to exploit local directional edge information, and a distribution field histogram \cite{dft} to balance descriptor specificity and the landscape smoothness criterion. In addition, local binary patterns (LBP) \cite{cxt}, scale-invariant feature transform (SIFT) \cite{sift} and Haar-like features \cite{ct} have also been explored to model object appearance. Handcrafted features only achieve partial success in coping with the challenges of appearance variations, scale change, and pose changes, yet requiring domain expertise for appropriate designs. On the other hand, automated feature extraction is able to learn self-taught features \cite{selftaught} from input images, which can be either unsupervised such as principal component analysis (PCA) \cite{ivt}, or supervised such as  linear discriminant analysis (LDA) \cite{lda}. For example, many recent tracking methods including local linear coding (LLC) \cite{llc}, sparsity-based collaborative model (SCM) \cite{scm}, multi-task tracker (MTT) \cite{mtmvt} and multi-view tracker (MVT) \cite{mvt} use learned sparse representations to improve the robustness against various target variations. However, due to the heavy computations involved in learning and optimization, most of these automated features trade real-time efficiency for the robustness. In summary, both handcrafted and automated features have their limitations, and it remains an important task to develop better feature representations of objects in visual tracking.

A tracking model is used to verify the prediction of any state, which can be generative or discriminative. In generative models \cite{sfa,simultaneous}, tracking is formulated as a search for the region within a neighborhood that is most similar to the target object. A variety of search methods based on generative models have been developed to estimate object states; for instance, a generalized Hough transform \cite{ht} for tracking-by-detection, a sparsity-based local appearance generative model \cite{asla}, and an object similarity metric with a gradient-based formulation \cite{dsat} to locate objects. However, generative models ignore negative samples in the background, resulting in vulnerability caused by background confusion. Recently, discriminative models \cite{discriminative,nonrigid} have been developed which consider the information of the object and the background simultaneously, and learn binary classifiers in an online manner to separate the foreground from the background. Numerous classifiers have been adopted for object tracking, such as multi-view SVM (MVS) \cite{msvm}, structured SVM (Struck) \cite{struck}, online AdaBoost (OAB) \cite{oab}, and online multi-instance learning (MIL) \cite{mil}, as well as several semi-supervised models \cite{dml}. However, discriminative models pay insufficient attention to the eigen basis of the tracking target and are unable to evaluate the credibility of the tracking results precisely. Therefore, a successful model should exploit the advantages of both generative and discriminative methods \cite{scm,cgdt} to account for appearance variations and to effectively separate the foreground target from the background.

Primates are acknowledged to be capable in high performance visual pattern recognition, as they can achieve invariance and discrimination uniformly. Recent research findings in brain cognition and computer vision demonstrate that the bio-inspired models are valuable in enhancing the performance of object recognition \cite{objectrecognition}, face identification \cite{faceidentification}, and scene classification \cite{sceneclassification}. We expect applying visual cortex research to object tracking would also be feasible and meaningful. Robust tracking target representation based on expert biological knowledge is able to avoid the parameter adjustment of handcrafted features and the parameter learning of automated features. In addition, the biological visual cognitive mechanism provides the inspiration for combining the generative model and discriminative model to handle appearance variations and to separate the target from the background effectively. This paper develops biologically inspired tracker (BIT) based on the ventral stream in the visual cortex. In line with expert biological knowledge and heuristics, a new bio-inspired appearance model is proposed which simulates multi-cue selective (classical simple cells, S1 units) and multi-variant competitive (cortical complex cells, C1 units) mechanism in shallow neurons to target representation, and achieves an appropriate trade-off between discrimination and invariance. A two-layer bio-inspired tracking model proposed for advanced learning combines the generative and discriminative model: the response of view-tuned learning (S2 units) is a generative model via convolution and a fully connected classifier simulates neuronal network for task-turned learning (C2 units) as a discriminative model. BIT exploits fast Gabor approximation (FGA) to speed up low-level bio-inspired feature extraction (S1 units and C1 units) and fast Fourier transform (FFT) to speed up high-level bio-inspired learning and dense sampling (S2 units and C2 units).

To evaluate the proposed BIT in terms of tracking accuracy, robustness, and efficiency, we conduct extensive experiments on the CVPR2013 tracking benchmark (TB2013) \cite{cvpr2013} and the Amsterdam Library of Ordinary Videos (ALOV300++) database \cite{alov}. Experimental results show  that BIT outperforms existing top-performing  algorithms in terms of accuracy and robustness. Moreover, BIT enhances speed via fast Gabor approximation and fast Fourier transform. It processes approximately 45 frames per second for object tracking on a computer equipped with Intel i7 3770 CPU (3.4GHz) and is therefore suitable for most real-time applications.

The remainder of this paper is organized as follows. In Section \ref{sec:Relatedwork}, we review the research works related to bio-inspired models and the tracking methods based on bio-inspired models. In Section \ref{sec::BIT}, we introduce the proposed biologically inspired tracker (BIT) and discuss both the appearance model and the tracking model in detail. Section \ref{sec:experiments} gives the experimental results on both the qualitative and quantitative analysis, including the comparison with other methods and an analysis of each part of our method. We conclude our paper in Section \ref{sec:conclusion}.

\section{Related Work}\label{sec:Relatedwork}
Humans and primates outperform the best machine vision systems on all vision tasks with regard to most measures, and thus it is critical yet attractive to emulate object tracking in the visual cortex. Understanding how the visual cortex recognizes objects is a critical question for neuroscience. A recent theory \cite{hierarchicalmodel} on the feed-forward path of visual processing in the cortex is based on the ventral stream processing from the primary visual cortex (V1) to the prefrontal cortex (PFC), which is modeled as a hierarchy of increasingly sophisticated representations. In the ventral stream, a bio-inspired model called HMAX \cite{hmax} has been successfully applied to machine vision and consists of alternating computational units called simple (S) and complex (C) cell units. In the primary visual cortex, the simple units combine intensity inputs with a bell-shaped tuning function to increase scale and direction selectivity. The complex units pool their inputs through a pooling operation (e.g. MAX, AVG or STD), thereby introducing gradual invariance to scale and translation. In the inferotemporal (IT) cortex (the so-called view-tuned units), samples of features were suggested that were highly selective for particular objects while being invariant to ranges of scales and positions. Note that a network comprised of units from the IT cortex to PFC is among the most powerful in terms of learning to generalize, which is equivalent to regularizing a classifier on its outputs tuning function.

Deep neural networks as a bio-inspired model can self-learn features from raw data without resorting to manual tweaking, and have achieved state-of-the-art results in several complicated tasks, such as image classification \cite{imagenetdcnn}, object recognition and segmentation \cite{rfh}. However, considerably less attention has been given to applying deep networks for visual tracking, because only the target state in the first frame is available to train deep networks. Fan et al. \cite{cnnhumantracking} proposed a human tracking algorithm that learns a specific feature extractor with convolutional neural networks from an offline training set of 20000 samples. In \cite{dlt}, Wang and Yeung proposed a deep learning tracking method that uses a stacked de-noising auto-encoder to learn the generic features from 1 million images sampled from the Tiny Images dataset \cite{80mti}. Both methods pay particular attention to the offline learning of an effective feature extractor with a large amount of auxiliary data, yet they do not fully take into account the similarities in local structural and inner geometric layout information between the targets over subsequent frames, which is useful and effective for distinguishing the target from the background for visual tracking. In addition, the above tracking methods based on deep networks cannot guarantee real-time performance because multilayer operations take intensive computation.

In the ventral stream, an HMAX model \cite{hmax} proposed for object recognition and scene classification has proven to be effective. However, the original bio-inspired model is designed for classification problems and is therefore difficult to apply straightforwardly to object tracking. In \cite{dst}, A bio-inspired tracker called discriminant saliency tracker (DST) uses a feature-based attention mechanism and a target-tuned top-down discriminant saliency detector for tracking, which does not explicitly retain appearance models in the previous frames resulting in invalidation in the challenges of occlusion or background cluster. Zheng et al. \cite{pso} presented an object tracking algorithm by using a bio-inspired population-based search algorithm, called particle swarm optimization (PSO). PSO introduces the cognitive mechanism of birds and fishes to accelerate the convergence in the potential solution space, so that the optimal solution can be found in a short time. However, PSO only focuses on the tracking model and loses sight of the importance of features. Li et al. \cite{sbif} proposed a simplified biologically inspired feature (SBIF) for object representation and combined SBIF with a Bayesian state inference tracking framework which utilized the particle filter to propagate sample distributions over time. The SBIF tracker extracts a robust representation feature, but ignores the advanced learning mechanism in the ventral stream and applies the time-consuming particle filter as the tracking model. In \cite{cnnt}, a simple convolutional network encodes the local structural information of the tracking object using a set of normalized patches as filters, which are randomly extracted from the target region in the first frame. All the convolution maps are extracted from raw intensity images, therefore no importance is attached to the low-level bio-inspired feature. In this paper, a novel visual tracker based on a bio-inspired model is developed  to improve the aforementioned  shortcomings.

\section{Biologically Inspired Tracker}\label{sec::BIT}
\begin{figure*}[!t]
\centering
\includegraphics[width=0.75\linewidth]{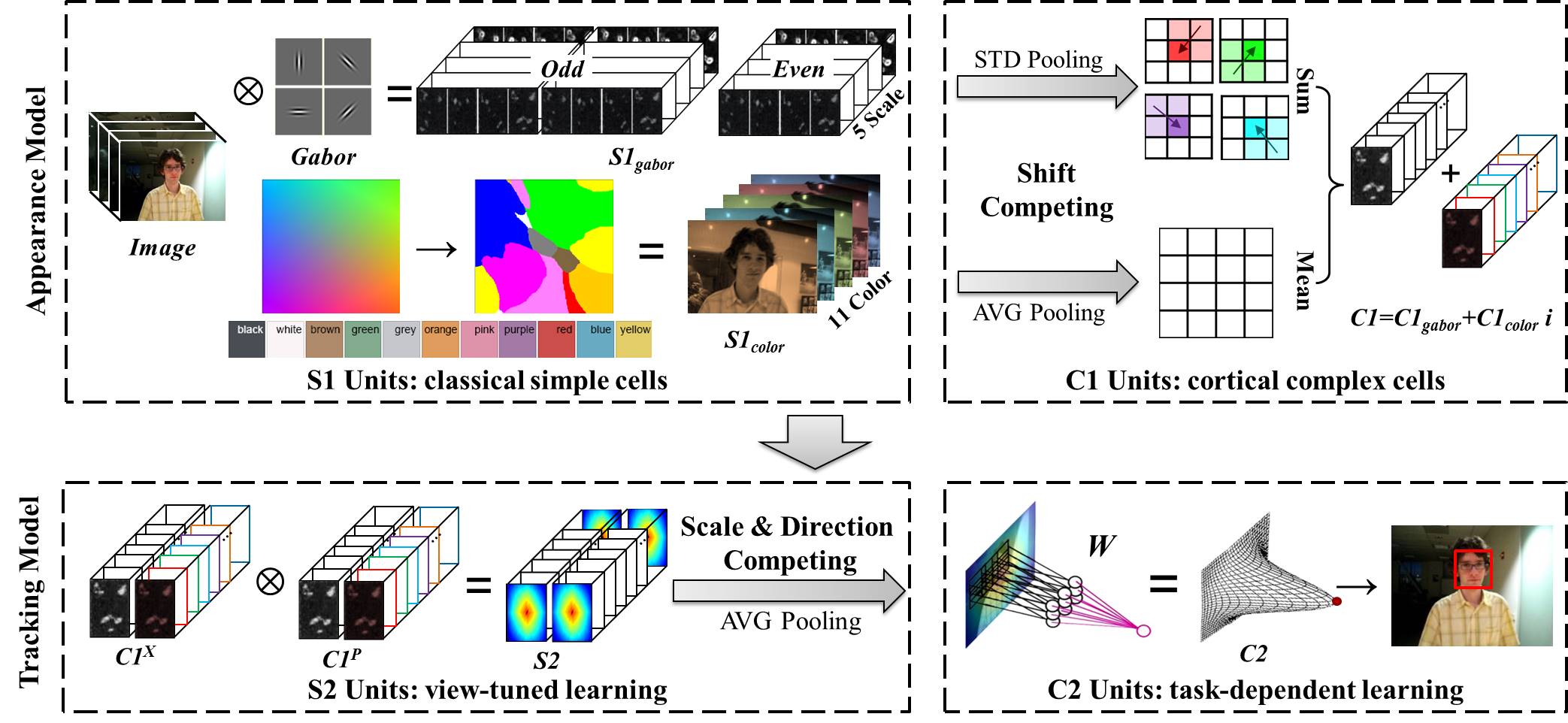}
\caption{Biologically inspired tracker. BIT cascades four units including appearance model (S1 and C1 units) and tracking model (S2 and C2 units): S1 units extract texture and color information by Gabor filters and color names; C1 units pool texture and color features and combine them by complex response maps; S2 units learn view-turned feature by a linear convolution between the input $X$ and the stored prototype $P$; C2 units apply a full-connection neural network for task-dependent learning.}
\label{fig:framework}
\end{figure*}

BIT is an HMAX model \cite{hmax} that partially mimics the ventral stream. It is particularly designed for visual tracking, and can be divided into two main components: a bio-inspired appearance model and a bio-inspired tracking model (as shown in Fig. \ref{fig:framework}). The bio-inspired appearance model is subdivided into classical simple cells (S1 units) and cortical complex cells (C1 units); the bio-inspired tracking model is divided into view-tuned learning (S2 units) and task-dependent learning (C2 units). In addition, FGA and FFT are exploited to significantly save computations.

\subsection{Bio-inspired Appearance Model}\label{sec:BIF}
Appearance representation plays an important role in a vision system, and we present a bio-inspired appearance model. In line with bioresearch on the striate cortex and extrastriate cortex, classical simple cells (S1 units) show that a variety of selective and cortical complex cells (C1 units) maintain feature invariance. The proposed bio-inspired appearance model achieves the unification of invariance and discrimination between different tracking targets by a hierarchical model.

\subsubsection{S1 units -- classical simple cells}\label{sec:S1}
In the primary visual cortex (V1) \cite{2dvcf}, a simple cell receptive field has the basic characteristics of multi-orientation, multi-scale and multi-frequency selection. S1 units can be described by a series of Gabor filters \cite{cortex-like}, which have been shown to provide an appropriate model of cortical simple cell receptive fields. In contrast to the traditional bio-inspired feature of the HMAX model, which uses single even Gabor filters (contrast insensitive), we consider that odd Gabor filters (contrast sensitive) are also important for target representation, because they can extract not only the texture intensity but also the texture gradient direction. The odd and even Gabor functions have been shown to provide a good model of cortical simple cell receptive fields and are defined by
\begin{equation}\label{gabor}
\left\{ \begin{array}{l}
{G_{even}}\left( {x,y,\theta ,s\left( {\sigma ,\lambda } \right)} \right) = \exp \left( { - \dfrac{{{X^2} + {\gamma ^2}{Y^2}}}{{2{\sigma ^2}}}} \right) \cos \left( {\dfrac{{2\pi }}{\lambda }X} \right)\\
{G_{odd}}\left( {x,y,\theta ,s\left( {\sigma ,\lambda } \right)} \right) = \exp \left( { - \dfrac{{{X^2} + {\gamma ^2}{Y^2}}}{{2{\sigma ^2}}}} \right) \sin \left( {\dfrac{{2\pi }}{\lambda }X} \right),
\end{array}\right.
\end{equation}
where $X = x\cos \theta  + y\sin \theta$, $Y =  - x\sin \theta  + y\cos \theta$, the filter patch coordinate $\left(x,y\right)$, the orientation $\theta$, scales $s$ with 2 parameters (the effective width $\sigma$ and the wavelength $\lambda$). Following \cite{cortex-like}, we arranged a series of Gabor filters to form a pyramid of scales, spanning a range of sizes from $7\times7$ to $15\times15$ pixels in steps of two pixels to model the receipt-field $\xi$ of the simple cells (parameter values of Gabor filters shown in Table \ref{tab:parameters} are the standard setting in bio-inspired model). The filters come in 4 orientations ($\theta  = 0,{\pi  \mathord{\left/
 {\vphantom {\pi  4}} \right.
 \kern-\nulldelimiterspace} 4},{\pi  \mathord{\left/
 {\vphantom {\pi  2}} \right.
 \kern-\nulldelimiterspace} 2},{{3\pi } \mathord{\left/
 {\vphantom {{3\pi } 4}} \right.
 \kern-\nulldelimiterspace} 4}$) on even Gabor filters and 8 orientations ($\theta  = 0, \pm {\pi  \mathord{\left/
 {\vphantom {\pi  4}} \right.
 \kern-\nulldelimiterspace} 4},{{ \pm \pi } \mathord{\left/
 {\vphantom {{ \pm \pi } 2}} \right.
 \kern-\nulldelimiterspace} 2},{{ \pm 3\pi } \mathord{\left/
 {\vphantom {{ \pm 3\pi } 4}} \right.
 \kern-\nulldelimiterspace} 4},\pi $) on odd Gabor filters, thus leading to 60 different S1 receptive field types in total. The corresponding result of classical simple cells is given by
\begin{equation}\label{S1gabor}
S{1_{gabor}}\left( {x,y,\theta ,s} \right) = I\left( {x,y} \right) \otimes G_{even/odd}\left( {x,y,\theta ,s} \right),
\end{equation}
where $I\left(x,y\right)$ is the original gray-scale image of tracking sequences.

In scene classification \cite{itti} and saliency detection \cite{saliency}, the joint color and texture information is shown to be important. To effectively represent a color target, we unify S1 units with both the color and texture information. The color units are inspired by the color double-opponent system in the cortex \cite{doubleopponent}. Neurons are fired by a color (e.g., blue) and inhibited by another color (e.g., yellow) in the center of the receptive field, as are neurons in the surrounding area. Color names (CN) \cite{cn} are employed to describe objects, and these are linguistic color labels assigned by humans to represent colors in the real world. There are 11 basic colors: black, brown, green, pink, red, yellow, blue, grey, orange, purple and white. However, the use of RGB color in computer vision can usually be mapped to a probabilistic 11 dimensional color attributes by the mapping matrix, which is automatically learned from images retrieved by Google Images search. Color name probabilities are defined by
\begin{equation}\label{S1color}
S1_{color}\left( {x,y,c} \right) = Map\left( {R\left( {x,y} \right),G\left( {x,y} \right),B\left( {x,y} \right),c} \right),
\end{equation}
where $R\left( {x,y} \right)$, $G\left( {x,y} \right)$, $B\left( {x,y} \right)$ correspond to the RGB color values of images, $c$ is the index of color names, and $Map()$ indicates a mapping matrix from RGB to 11 dimensional color probabilities. In order to keep the same feature dimension as $S1_{gabor}$ in one scale, $S1_{color}\left(x,y,12\right)$ is set to 0 and $S1_{color}$ is set to 0 for gray image.

In this paper, a color target is represented by 60 feature maps of complex number, of which 60 texture feature maps are obtained from multi-scale Gabor filters convoluted with the input image as the real part, and 12 color feature maps are copied five times corresponding to 5 scales as the imaginary part. Combining the texture feature with the color feature through the real and imaginary parts, complex feature maps maintain the balance between different types of features and take full advantage of the complex frequency signal information of FFT to reduce the computation operations by almost half.

\subsubsection{C1 units -- cortical complex cells}\label{sec:C1}

The cortical complex cells (V2) receive the response from simple cells and have the function of primary linear feature integration. C1 units \cite{catsfields} correspond to complex cells, which show the invariance to have larger receptive fields (shift invariance). Ilan et al. \cite{maxpool} suggested that the spatial integration properties of complex cells can be described by a series of pooling operations. Riesenhuber and Poggio \cite{hierarchicalmodel} argued for and demonstrated the advantages of using the nonlinear MAX operator over the linear summation operation SUM, and Guo et al. \cite{age} proposed another nonlinear operation called standard deviation STD for human age estimation. In this paper, a different pooling method is used for shift competing on each different feature map.

In order to keep C1 unit features what are changes in bias, invariance to gain can be achieved via local normalization. In \cite{age}, the STD operation has been shown to outperform pure MAX pooling for revealing the local variations that might be significant for characterizing subtlety. Dalal and Triggs \cite{hog} used four different normalization factors for the histograms of oriented gradients. Based on \cite{hog,age}, an enhanced STD operation in a cell grid $\Sigma$ of size $n_s\times n_s=4\times4$ is shown as \eqref{C1gabor}, where ${N_{{\delta _x},{\delta _y}}}\left( {x,y} \right)$ is a normalization factor and $\left( {{\delta _x},{\delta _y}} \right)$ is the shift bias.
\begin{equation}\label{C1gabor}
C1_{gabor}\left( {x,y} \right) = \sum\limits_{\scriptstyle\left( {x,y} \right) \in \Sigma \hfill\atop
\scriptstyle{\delta _x},{\delta _y} \in  \pm 1\hfill} {\dfrac{{S1_{gabor}\left( {x,y} \right)}}{{{N_{{\delta _x},{\delta _y}}}\left( {x,y} \right)}}}
\end{equation}
\begin{equation}\label{Nxy}
\begin{array}{l}
{N_{{\delta _x},{\delta _y}}}\left( {x,y} \right) = (S1_{gabor}^2\left( {x,y} \right) + S1_{gabor}^2\left( {x + {\delta _x},y + {\delta _y}} \right)\\
\quad \quad \quad \quad + S1_{gabor}^2\left( {x + {\delta _x},y} \right) + S1_{gabor}^2\left( {x,y + {\delta _y}} \right){)^{0.5}}
\end{array}
\end{equation}

In addition, the mapping of the S1 color units from the RGB image by point-to-point is sensitive to noise, therefore an AVG pooling operation is used for color features. The C1 color unit responses are computed by subsampling these maps using a cell grid $\Sigma$ of size $n_s\times n_s=4\times4$ as C1 texture units. From each grid cell, a single measurement is obtained by taking the mean of all 16 elements. In summary, $C1_{color}$ response is
\begin{equation}\label{C1color}
C1_{color}\left( {x,y} \right) = \frac{1}{{{n_s} \times {n_s}}}\sum\limits_{\left( {x,y} \right) \in \Sigma } {S1_{color}\left( {x,y} \right)}
\end{equation}

\subsection{Bio-inspired Tracking Model}\label{sec:BIM}
The tracking model is developed to verify any object state prediction, which can be either generative or discriminative. Based on the learning mechanism of advanced neurons, the IT cortex and PFC are explored to design bio-inspired tracking model. In this paper, a tracking method based on an advanced learning mechanism is presented to combine generative and discriminative models, corresponding to the view-tuned learning (S2 units) and the task-dependent learning (C2 units).

\subsubsection{S2 units -- view-tuned learning}\label{sec:S2}
The tuning properties of neurons in the ventral stream of the visual cortex, from V2 to the IT cortex, play a key role in visual perception in primates, and in particular for object recognition abilities \cite{bio}. This training process can be regarded as a generative model, in which S2 units pool over afferent C1 units within its receptive field. Each S2 unit response depends on a radial basis function (RBF) \cite{cortex-like} on the Euclidean distance between a new input $X$ and a stored prototype $P$. For an image patch from the previous C1 layer, the response $r_{S2}$ of the corresponding to S2 units is given by:
\begin{equation}\label{rs2}
{r_{S2}} = \exp \left( { - \beta {{\left\| {X - P} \right\|}^2}} \right),
\end{equation}
where $\beta$ defines the sharpness of the tuning coefficient. At runtime, S2 response maps are computed across all positions by \eqref{rs2} for each band of C2 units.

According to \eqref{rs2}, we know the S2 units response corresponds to a kernel method based on RBF, which can be rewritten similarly to the linear function as follows \eqref{rs2l}, when the RBF is a standard normal function ($\beta  = {1 \mathord{\left/
 {\vphantom {1 {2{\sigma ^2}}}} \right.
 \kern-\nulldelimiterspace} {2{\sigma ^2}}} = {1 \mathord{\left/
 {\vphantom {1 2}} \right.
 \kern-\nulldelimiterspace} 2}$).
\begin{equation}\label{rs2l}
\begin{array}{l}
{r_{S2}} = \exp \left( { - \dfrac{1}{{2{\sigma ^2}}}{{\left\| {X - P} \right\|}^2}} \right)\\
\quad  = \exp \left( { - \dfrac{1}{2}\left( {{X^T}X + {P^T}P - 2{X^T}P} \right)} \right)\\
\quad  \sim \exp \left( {{X^T}P} \right) \sim {X^T}P
\end{array}
\end{equation}
Here $X^TX$ and $P^TP$ as self-correlation coefficient almost keep changeless nearly constant with marginal effects on S2 units, and $X^TP$ roughly locates in the linear region of exponential function. Moreover, linear kernel is usually preferred in time-critical problems such as tracking, because the weight vector can be computed explicitly. Therefore, the S2 units dense response map was calculated using a linear function instead of RBF. The C1 units response of new input $X$ is expressed as $C{1^X}\left( {x,y,k} \right)$ and the response of stored prototype $P$ is $C{1^P}\left( {x,y,k} \right)$, where $k$ is the index of 60 feature maps corresponding to 12 orientations and 5 scales. To achieve scale and orientation invariance, an AVG pooling operation is used for the fusion of multi-feature maps:
\begin{equation}\label{S2}
S2\left( {x,y} \right) = \frac{1}{K}\sum\limits_{k = 1}^K {C{1^X}\left( {x,y,k} \right) \otimes C{1^P}\left( {x,y,k} \right)}
\end{equation}
\subsubsection{C2 units -- task-dependent learning}\label{sec:C2}
The task-specific circuits from the IT cortex to the PFC learn the discrimination between target objects and background clusters. According to bioresearch \cite{bio}, the routine running in PFC as a classifier is trained on a particular task in a supervised way and receives the activity of a few hundred neurons in the IT cortex. The classifier can indeed read-out the object identity and object information (such as position and size of the object) from the activity of about 100 neurons in the IT cortex with high accuracy and in an extremely short time. Supervised learning at this stage involves adjusting the synaptic weights to minimize error in the training set. In this paper, a convolutional neural network (CNN) is used as \eqref{C2}, which corresponds to the task-specific circuits found in C2 units with neurons from the IT cortex to the PFC.
\begin{equation}\label{C2}
C2\left( {x,y} \right) = W\left( {x,y} \right) \otimes S2(x,y),
\end{equation}
where $W$ is the synaptic weights of the neural network. In addition, a fast estimate method of $W$ will be introduced in the next subsection \ref{sec:realBIT}.

\subsection{Real-time bio-inspired tracker}\label{sec:realBIT}
Due to the time-sensitive nature of tracking, modern trackers walk a fine line between incorporating as many samples as possible and keeping computational demand low. A practical visual tracker should reach a speed of at least 24 frames per second (the standard frame rate of films). Past visual trackers based on bio-inspired models are unable to maintain real-time because of complex neuron simulation, such as SBIF tracker \cite{sbif} and discriminant saliency tracker \cite{dst}. In this paper, an FGA and an FFT are applied to speed up the hierarchical bio-inspired model.

\subsubsection{Fast Gabor Approximation (FGA)}\label{sec:FGA}
To retain the invariance of bio-inspired feature, a series of multi-scale and multi-orientation Gabor filters is used for convolution with gray images. Therefore, a 60 times convolution operation seriously affects the instantaneity of the BIT framework. In this paper, we propose FGA to reduce the computational cost, which is inspired by histograms of oriented gradients (HOG)\cite{hog}. HOG uses horizontal and vertical gradients to approximate the gradient intensity in each orientation. Using several pairs of 1-D Gabor filters (${G_x}\left( {x,s\left( {\sigma ,\lambda } \right)} \right)$ and ${G_y}\left( {y,s\left( {\sigma ,\lambda } \right)} \right)$), which are at 5 different scales and orthogonal to each other as in Section \ref{sec:S1}, 10 multi-scale orthogonal Gabor response maps at a pixel $(x,y)$ are computed as
\begin{equation}\label{Gxy}
\begin{array}{l}
\left\{ \begin{array}{l}
{D_x}\left( {x,y,s\left( {\sigma ,\lambda } \right)} \right) = I\left( {x,y} \right) \otimes {G_x}\left( {x,s\left( {\sigma ,\lambda } \right)} \right)\\
{D_y}\left( {x,y,s\left( {\sigma ,\lambda } \right)} \right) = I\left( {x,y} \right) \otimes {G_y}\left( {y,s\left( {\sigma ,\lambda } \right)} \right)
\end{array} \right.\\
where~\left\{ \begin{array}{l}
{G_x}\left( {x,s\left( {\sigma ,\lambda } \right)} \right) = \exp \left( { - \dfrac{{{x^2}}}{{2{\sigma ^2}}}} \right) \times \sin \left( {\dfrac{{2\pi }}{\lambda }x} \right)\\
{G_y}\left( {y,s\left( {\sigma ,\lambda } \right)} \right) = \exp \left( { - \dfrac{{{y^2}}}{{2{\sigma ^2}}}} \right) \times \sin \left( {\dfrac{{2\pi }}{\lambda }y} \right)
\end{array} \right.
\end{array}
\end{equation}
Let $\Theta \left( {x,y,s\left( {\sigma ,\lambda } \right)} \right)$ and $A\left( {x,y,s\left( {\sigma ,\lambda } \right)} \right)$ be the orientation and magnitude of the Gabor gradient at a pixel $(x,y)$ in an image showing as \eqref{atheta}, by which the response of the multi-orientation S1 units is approximated.
\begin{equation}\label{atheta}
\left\{ \begin{array}{l}
\Theta \left( {x,y,s\left( {\sigma ,\lambda } \right)} \right) = {\tan ^{ - 1}}\left( {\dfrac{{{D_y}\left( {x,y,s\left( {\sigma ,\lambda } \right)} \right)}}{{{D_x}\left( {x,y,s\left( {\sigma ,\lambda } \right)} \right)}}} \right)\\
A\left( {x,y,s\left( {\sigma ,\lambda } \right)} \right) = \sqrt {D_x^2\left( {x,y,s} \right) + D_y^2\left( {x,y,s} \right)}
\end{array} \right.
\end{equation}
We define a pixel-level feature map that specifies a sparse response of Gabor magnitudes at each pixel $(x,y)$ to approximate multi-orientation in the S1 units. When $\Theta\left({x,y,s\left( {\sigma ,\lambda } \right)}\right)$ belongs to the range of corresponding orientations, the magnitude $A\left({x,y,\theta ,s\left( {\sigma ,\lambda } \right)}\right)$ is set as the approximate response of S1 units as
\begin{equation}\label{S1odd}
S{1_{odd}}\left(\cdot \right) = \left\{ \begin{array}{l}
A\left(  \cdot  \right),if~ \Theta \left(  \cdot  \right) \in \left[ {\theta  - \pi/8,\theta  + \pi/8} \right)\;\\
0,otherwise
\end{array} \right.
\end{equation}
\begin{equation}\label{S1even}
S{1_{even}}\left(\cdot \right) = \left\{ \begin{array}{l}
A\left(  \cdot  \right),if~ \Theta \left(  \cdot  \right) \in \left[ {\theta  - \pi/8,\theta  + \pi/8} \right) \cup \\
\quad \quad \quad \quad \quad \quad \quad \left[ {\theta  + 7\pi/8,\theta  + 9\pi/8} \right)\;\\
0,otherwise
\end{array} \right.
\end{equation}

\subsubsection{Fast Fourier Transform (FFT)}\label{sec:FFT}
Many tracking approaches \cite{tld,scm,asla} have featured tracking-by-detection, which stems directly from the development of discriminative methods in machine learning. Almost all of the proposed tracking-by-detection methods are based on a sparse sampling strategy. In each frame, a small number of samples are collected in the target neighborhood by particle filter, because the cost of not doing so would be prohibitive. Therefore, speeding up the dense sampling of the S2 and C2 response calculation is a key feature of BIT. In this subsection, a real-time BIT based on dense sampling via FFT \cite{fft} will be introduced. At time $t$, a S2 units dense response map was calculated by a linear function instead of RBF as
\begin{equation}\label{s2t}
S{2_{t + 1}}\left( {x,y} \right) = \frac{1}{K}\sum\limits_{k = 1}^K {C1_{t + 1}^X\left( {x,y,k} \right) \otimes C1_t^P\left( {x,y,k} \right)}
\end{equation}
According to dot-product in frequency-domain equivalent to convolution in time-domain \cite{signals}, we note that \eqref{s2t} can be transformed to the frequency domain, in which FFT can be used for fast convolution. That is,
\begin{equation}\label{fs2t}
\mathcal{F}\left[ {S{2_{t + 1}}\left( {\cdot} \right)} \right] = \frac{1}{K}\sum\limits_{k = 1}^K {\mathcal{F}\left[ {C1_{t + 1}^X\left( {\cdot,k} \right)} \right] \odot \mathcal{F}\left[ {C1_t^P\left( {\cdot,k} \right)} \right]},
\end{equation}
where $\mathcal{F}\left[~\right]$ denotes the FFT function and $\odot$ is the elementwise dot-product.

As with S2 units, the FFT algorithm can also be used for fast convolution and deconvolution in C2 units. Note that the convolutional neural network is comprised of units with Gaussian-like tuning function together on their outputs. In order to estimate the neuronal connection weights W, the C2 units response map of an object location is modeled as
\begin{equation}\label{gaussian}
\widetilde{C2}\left( {x,y} \right) = \exp \left( { - \frac{1}{{2{\sigma_{s} ^2}}}\left( {{{\left( {x - {x_o}} \right)}^2} + {{\left( {y - {y_o}} \right)}^2}} \right)} \right),
\end{equation}
where $\sigma_{s}$ is a scale parameter and $\left({x_o},{y_o}\right)$ is the center of the tracking target. The neuron connection weights $W$ is therefore shown as
\begin{equation}\label{wt}
\mathcal{F}\left[ {W\left( x,y \right)} \right] =  \frac{{\mathcal{F}\left[ {\widetilde{C2}\left( {x,y} \right)} \right]}}{{\mathcal{F}\left[ {S2\left( {x,y} \right)} \right]}}
\end{equation}
The object location $\left( {\hat x,\hat y} \right)$ in the ($t+1$)-th frame is determined by maximizing the new C2 response map.
\begin{equation}\label{target}
\left( {\hat x,\hat y} \right) = \mathop {\arg \max }\limits_{\left( {x,y} \right)} C{2_{t + 1}}\left( {x,y} \right),
\end{equation}
where $C{2_{t + 1}}\left( {x,y} \right) = {\mathcal{F}^{ - 1}}\left[ {\mathcal{F}\left[ {{W_t\left( x,y \right)}} \right]\odot\mathcal{F}\left[ {S{2_{t + 1}}\left( {x,y} \right)} \right]} \right]$ and $\mathcal{F}^{ - 1}[~]$ denotes the inverse FFT function.

Depending on the spatial and frequency domains, a classical method of tracking model update is used in this paper. At the $t$-th frame, the BIT is updated by
\begin{equation}\label{update}
\left\{ \begin{array}{l}
C1_{t + 1}^P\left( { x,y,k } \right) = \rho C1\left( {\hat x,\hat y,k } \right) + \left( {1 - \rho } \right)C1_t^P\left( { x,y,k } \right)\\
\mathcal{F}\left[ {{W_{t + 1}}\left(  x,y  \right)} \right] = \rho \mathcal{F}\left[ {W\left( {\hat x,\hat y} \right)} \right] + \left( {1 - \rho } \right)\mathcal{F}\left[ {{W_t}\left(  x,y  \right)} \right]
\end{array} \right.,
\end{equation}
where $\rho$ is a learning parameter, $C1\left( {\hat x,\hat y, k } \right)$ is the C1 units spatial model and $\mathcal{F}\left[ {W\left( {\hat x,\hat y} \right)} \right]$ is the frequency model of neural weights computed by \eqref{wt}.

Based on FGA and FFT, the proposed real-time bio-inspired tracker is summarized in Algorithm \ref{alg:Tracking}.
\begin{algorithm}
\caption{Real-time Bio-inspired Tracker}
\begin{algorithmic}[1]\label{alg:Tracking}
\REQUIRE Gray images $I\left(x,y\right)$, Color images $RGB\left(x,y\right)$
\ENSURE Tracking result $\left( {\hat x,\hat y} \right)$
\STATE Set $\widetilde{C2}=\mathcal{F}\left(\exp \left( { - \dfrac{1}{{2{\sigma_{s} ^2}}}\left( {{{\left( {x - {x_o}} \right)}^2} + {{\left( {y - {y_o}} \right)}^2}} \right)} \right)\right)$
\STATE Set $C1^P\left( { x,y,k } \right) = C1\left( {x,y,k } \right)$ when $t=1$
\FOR{$t=1,2,\ldots$}
    \STATE \textbf{Bio-inspired appearance model}
    \STATE S1 units: \\
    $S1_{gabor}$ is calculated from $I\left(x,y\right)$ by \eqref{S1odd} and \eqref{S1even}.\\
    $S1_{color}$ is calculated from $RGB\left(x,y\right)$ by \eqref{S1color}.
    \STATE C1 units: \\
    $C{1_{gabor}}\left( {x,y,k} \right)$ is STD pooled from $S1_{gabor}$ by \eqref{C1gabor}.\\
    $C1_{color}\left( {x,y,k} \right)$ is AVG pooled from $S1_{color}$ by \eqref{C1color}.\\
    $C1\left(x,y,k\right)=C1_{gabor}\left(x,y,k\right)+C1_{color}\left(x,y,k\right)i$\\
    \STATE \textbf{Bio-inspired tracking model}
    \STATE S2 units:\\
    $\mathcal{F}\left[ {S2\left( {x,y} \right)} \right]=\dfrac{1}{K}\sum\limits_{k = 1}^K {\mathcal{F}\left[ {C1\left( {x,y,k} \right)} \right] \mathcal{F}\left[ {C1^P\left( {x,y,k} \right)} \right]}$\\
    \STATE C2 units:\\
    $C2\left( {x,y} \right) = {\mathcal{F}^{ - 1}}\left[ {\mathcal{F}\left[ {{W\left( x,y \right)}} \right]\odot\mathcal{F}\left[ {S2\left( {x,y} \right)} \right]} \right]$
    \STATE Find target: $\left( {\hat x,\hat y} \right) = \mathop {\arg \max }\limits_{\left( {x,y} \right)} C2\left( {x,y} \right)$\\
    \STATE Update Model:\\
    $C1^P\left( { x,y,k } \right) = \rho C1\left( {\hat x,\hat y,k } \right) + \left( {1 - \rho } \right)C1^P\left( { x,y,k } \right)$\\
    $\mathcal{F}\left[ {{W}\left(  x,y  \right)} \right] = \rho \mathcal{F}\left[ {W\left( {\hat x,\hat y} \right)} \right] + \left( {1 - \rho } \right)\mathcal{F}\left[ {{W}\left(  x,y  \right)} \right]$
\ENDFOR
\end{algorithmic}
\end{algorithm}

\section{Experiments}\label{sec:experiments}
We evaluate our method on two popularly used visual tracking benchmarks, namely the CVPR2013 Visual Tracker Benchmark (TB2013) \cite{cvpr2013} and the Amsterdam Library of Ordinary Videos Benchmark (ALOV300++) \cite{alov}. These two benchmarks contain more than 350 sequences and cover almost all challenging scenarios such as scale change, illumination change, occlusion, cluttered background, and/or motion blur. Furthermore, these two benchmarks evaluate tracking algorithms with different measures and criteria, which can be used to analyze the tracker from different views.

We use the same parameter values of BIT on the two benchmarks. Parameters of the bio-inspired appearance model are given in Table \ref{tab:parameters}. Tracking model parameters mentioned in Sec. \ref{sec:BIM} are specified as follows: the learning rate $\rho$ is set to 0.02, and the scale parameter $\sigma_{s}$ of $\widetilde{C2}$ is set to 0.1 or 0.08 according to the C2 response in the first five frames. (When the average trend is ascending, $\sigma_{s}$ is set to 0.1 or is set to 0.08 otherwise). The proposed tracker is implemented in MATLAB 2014A on a PC with Intel i7 3770 CPU (3.4GHz), and runs more than 45 frames per second (fps) on this platform.
\begin{table}
\center
\footnotesize
\caption{Summary of parameters used in S1 units}
\begin{tabular}{c|c|c|c}
\hline\hline
\textbf{Scale} $s$ & \textbf{Receipt-field} $\xi$ & $\sigma$ & $\lambda$\\
\hline
1&$7\times7$&2.8&3.5\\\hline
2&$9\times9$&3.6&4.6\\\hline
3&$11\times11$&4.5&5.6\\\hline
4&$13\times13$&5.4&6.8\\\hline
5&$15\times15$&6.3&7.9\\\hline
\hline
\end{tabular}
\label{tab:parameters}
\end{table}

\subsection{Comparison of results on the CVPR2013 benchmark}\label{sec:CVPR2013}
\begin{figure*}
  \centering
\includegraphics[width=0.094\linewidth,angle=0]{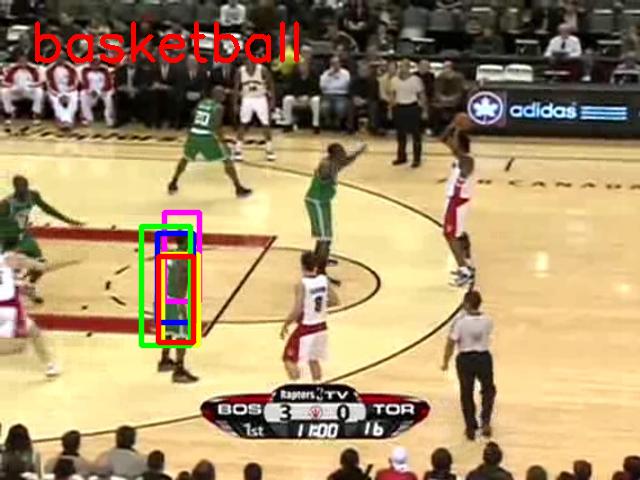}
\includegraphics[width=0.094\linewidth,angle=0]{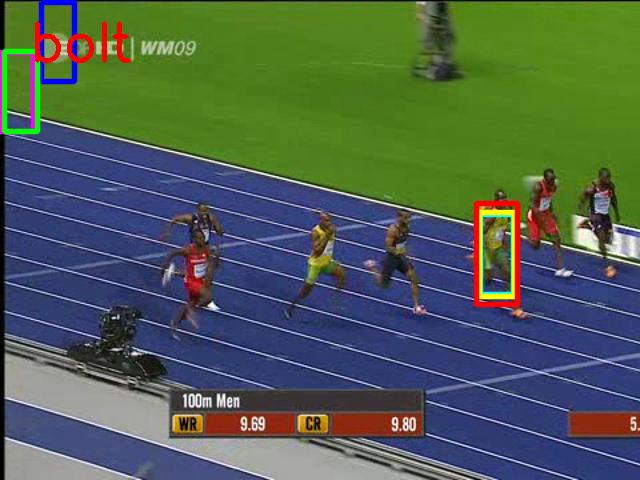}
\includegraphics[width=0.094\linewidth,angle=0]{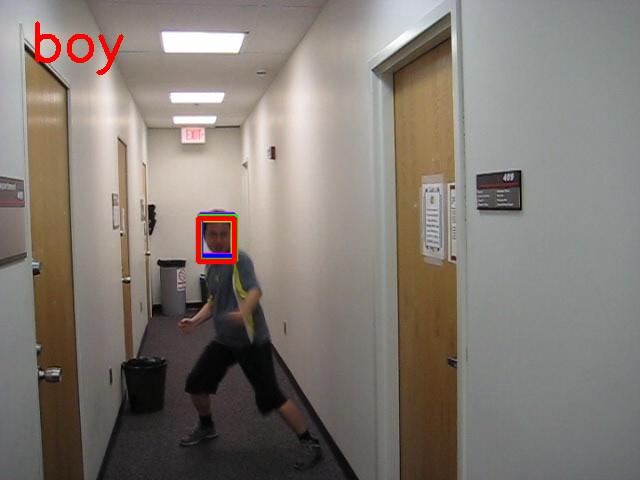}
\includegraphics[width=0.094\linewidth,angle=0]{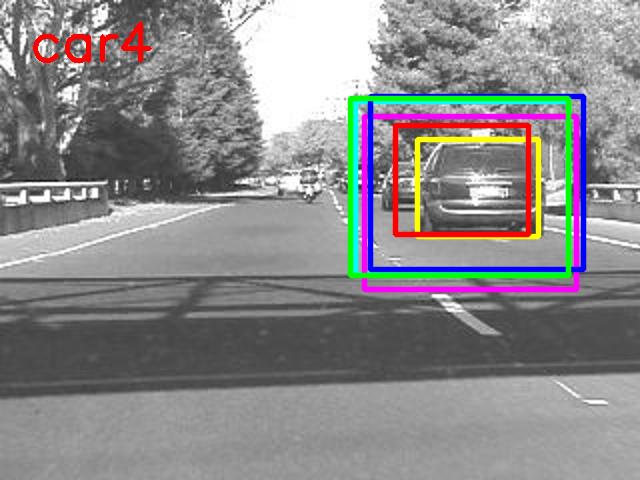}
\includegraphics[width=0.094\linewidth,angle=0]{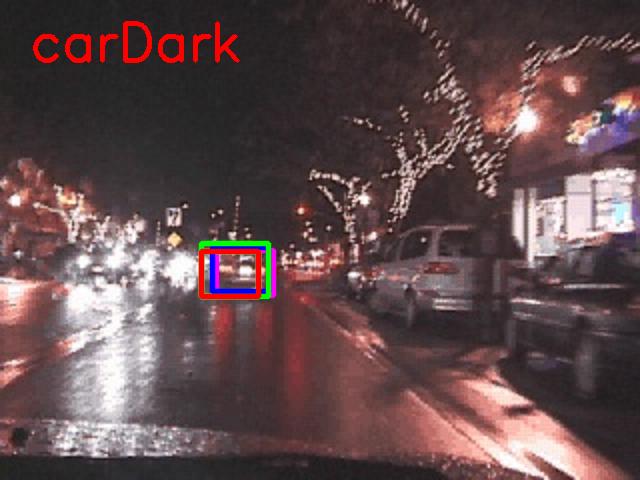}
\includegraphics[width=0.094\linewidth,angle=0]{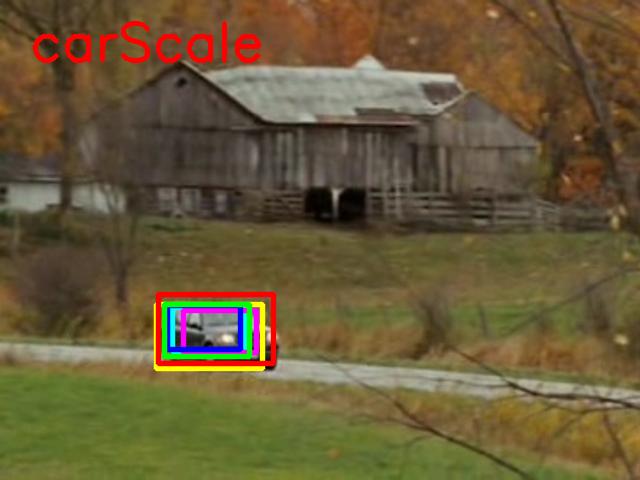}
\includegraphics[width=0.094\linewidth,angle=0]{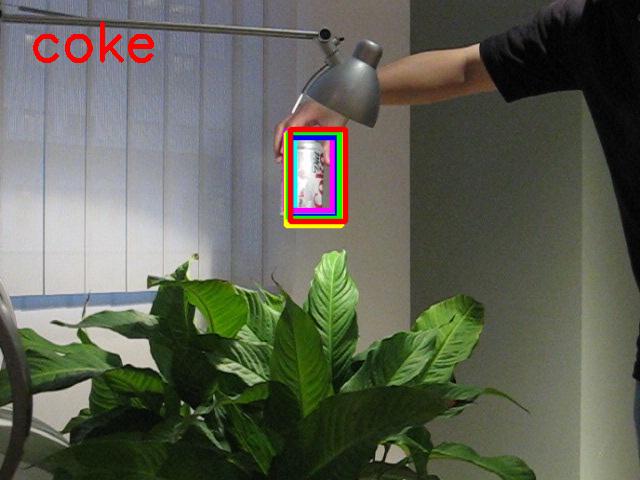}
\includegraphics[width=0.094\linewidth,angle=0]{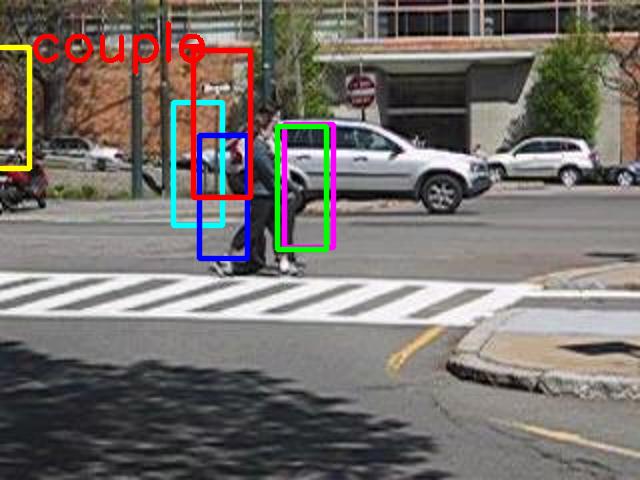}
\includegraphics[width=0.094\linewidth,angle=0]{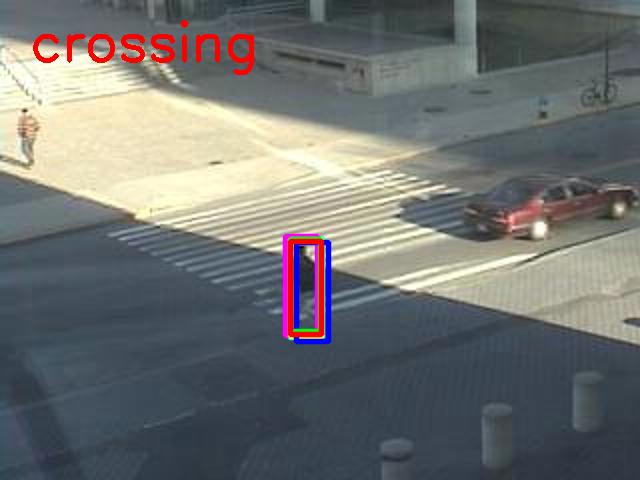}
\includegraphics[width=0.094\linewidth,angle=0]{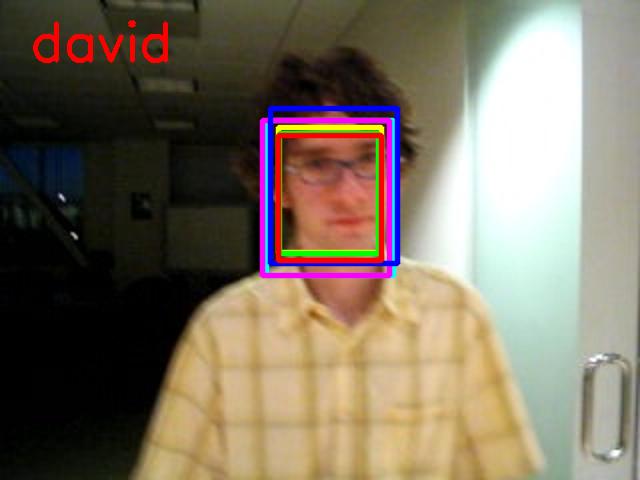}\\
\includegraphics[width=0.094\linewidth,angle=0]{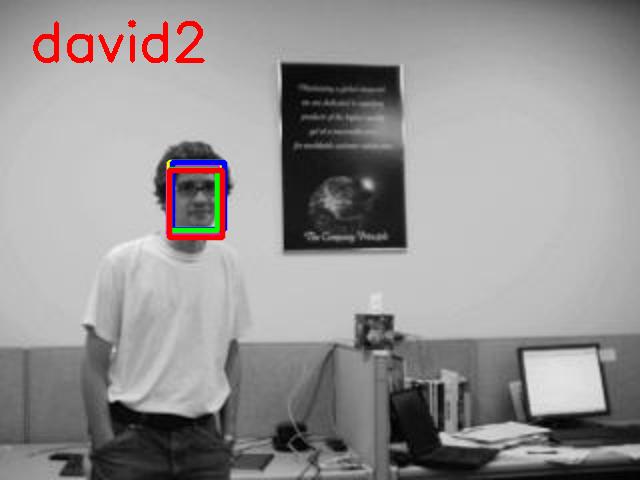}
\includegraphics[width=0.094\linewidth,angle=0]{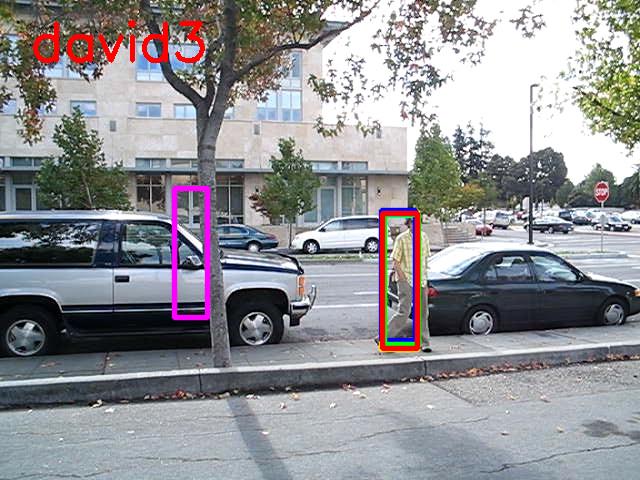}
\includegraphics[width=0.094\linewidth,angle=0]{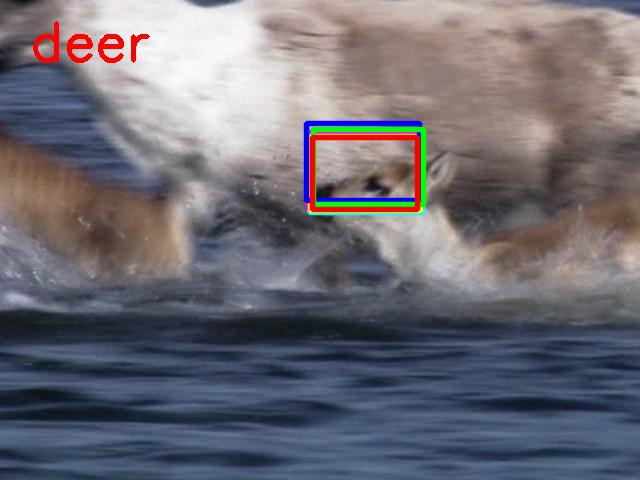}
\includegraphics[width=0.094\linewidth,angle=0]{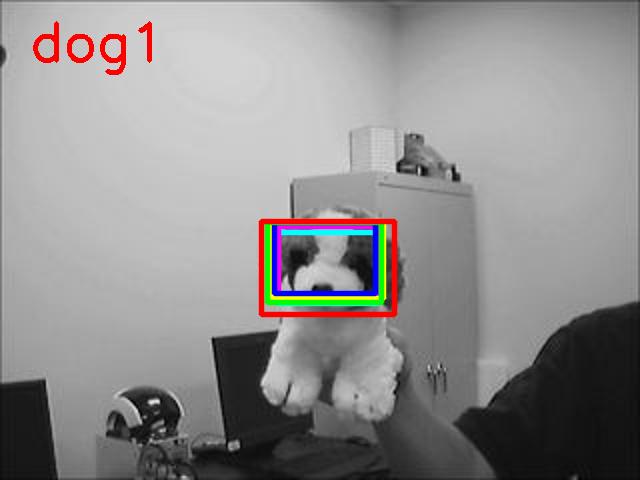}
\includegraphics[width=0.094\linewidth,angle=0]{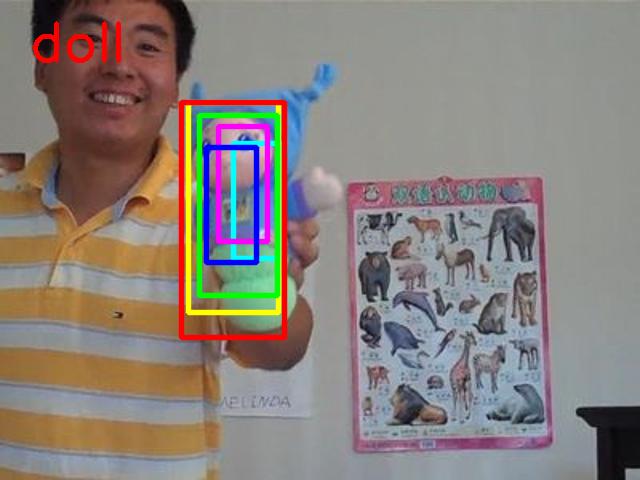}
\includegraphics[width=0.094\linewidth,angle=0]{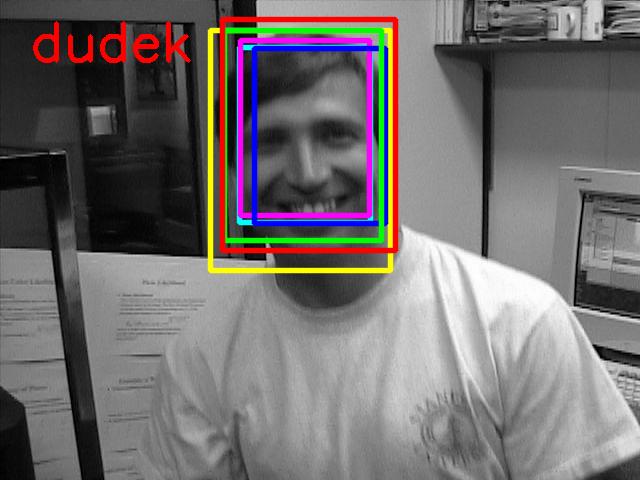}
\includegraphics[width=0.094\linewidth,angle=0]{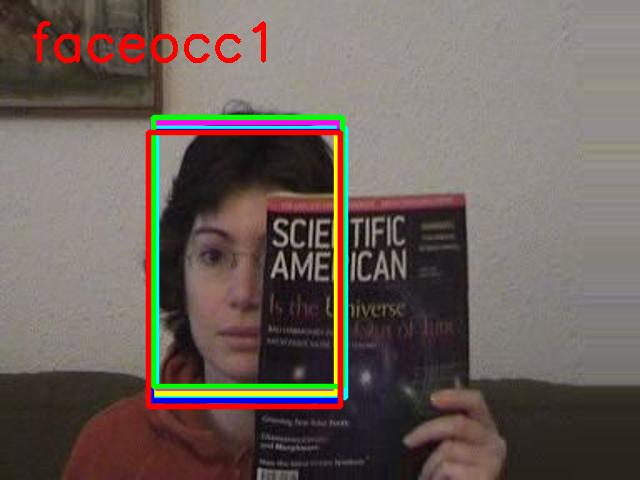}
\includegraphics[width=0.094\linewidth,angle=0]{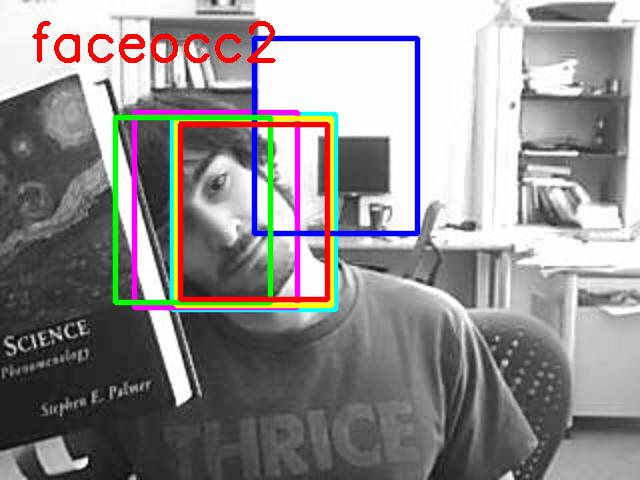}
\includegraphics[width=0.094\linewidth,angle=0]{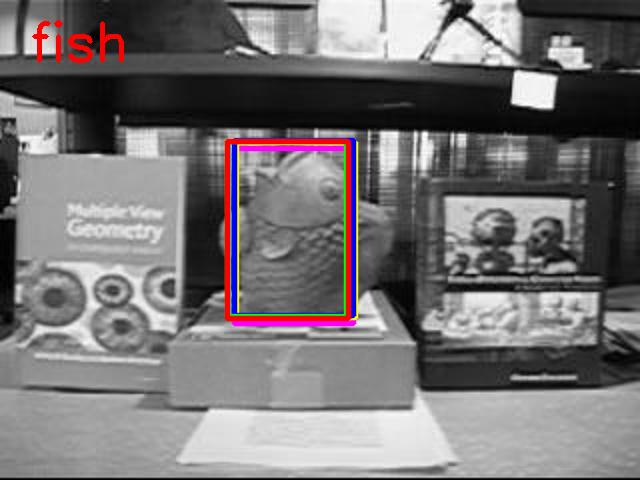}
\includegraphics[width=0.094\linewidth,angle=0]{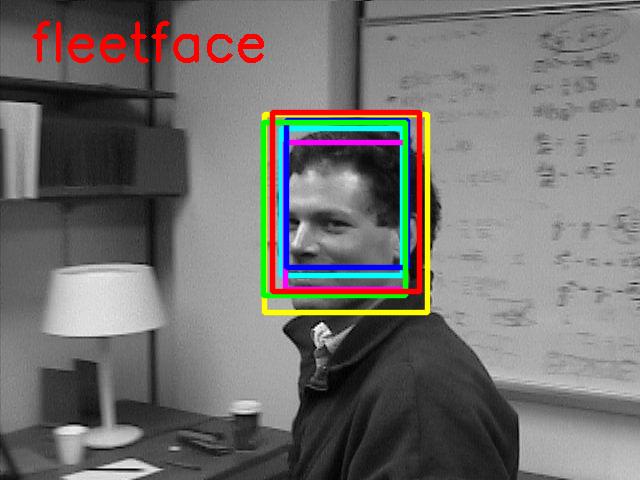}\\
\includegraphics[width=0.094\linewidth,angle=0]{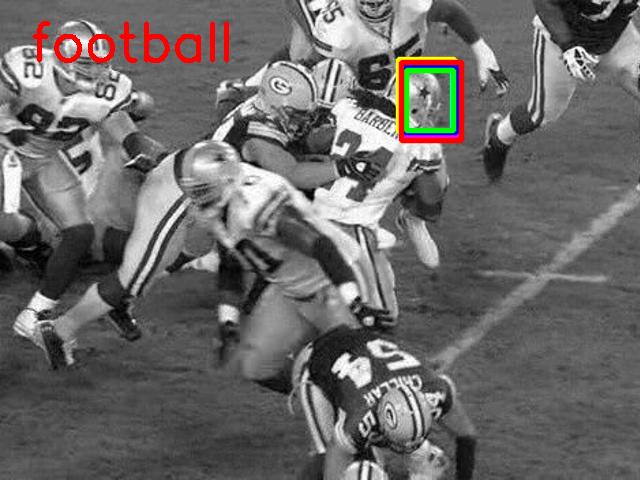}
\includegraphics[width=0.094\linewidth,angle=0]{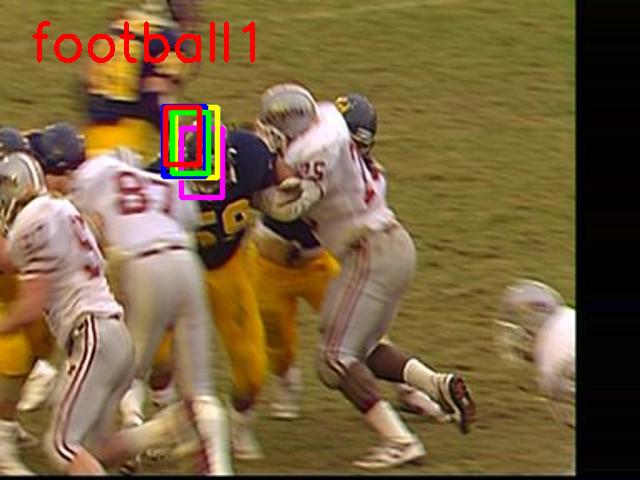}
\includegraphics[width=0.094\linewidth,angle=0]{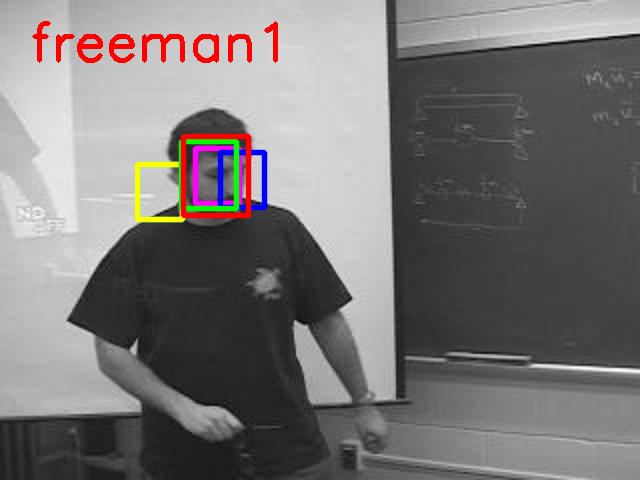}
\includegraphics[width=0.094\linewidth,angle=0]{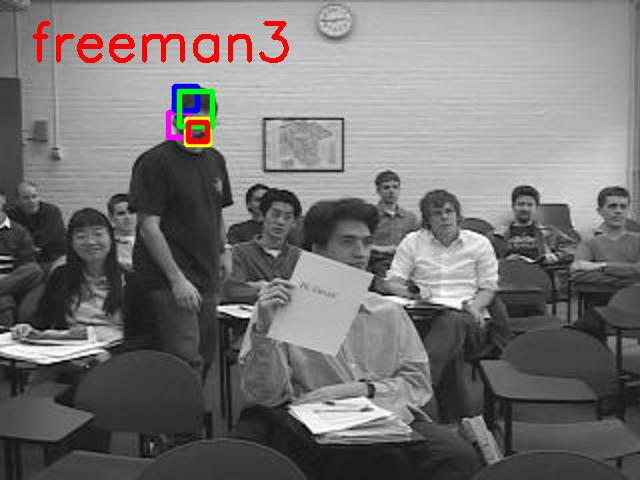}
\includegraphics[width=0.094\linewidth,angle=0]{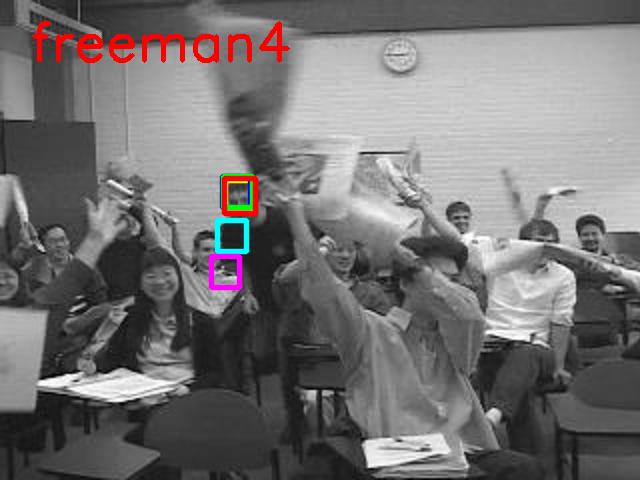}
\includegraphics[width=0.094\linewidth,angle=0]{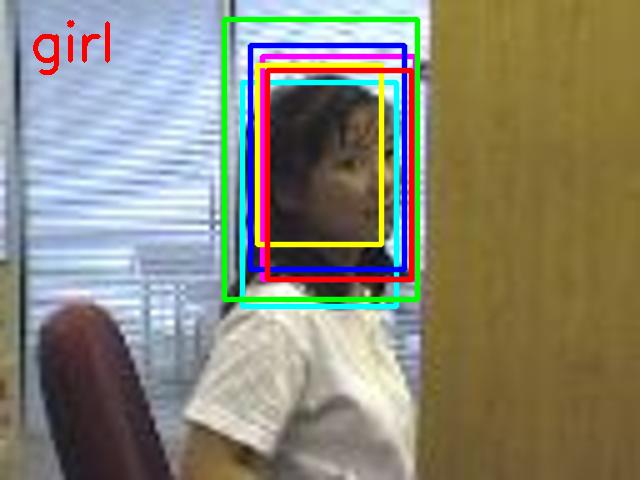}
\includegraphics[width=0.094\linewidth,angle=0]{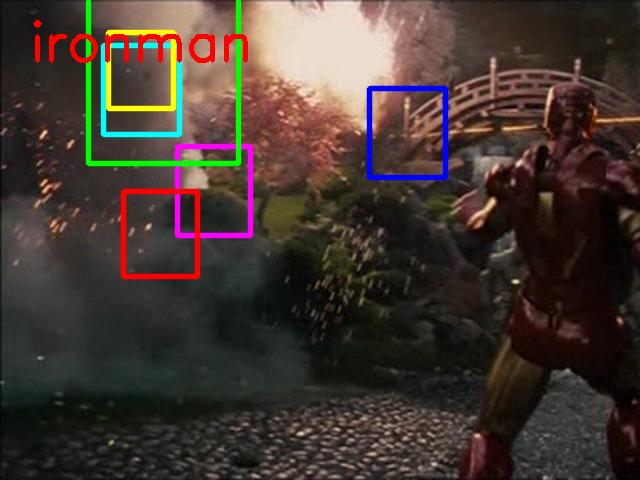}
\includegraphics[width=0.094\linewidth,angle=0]{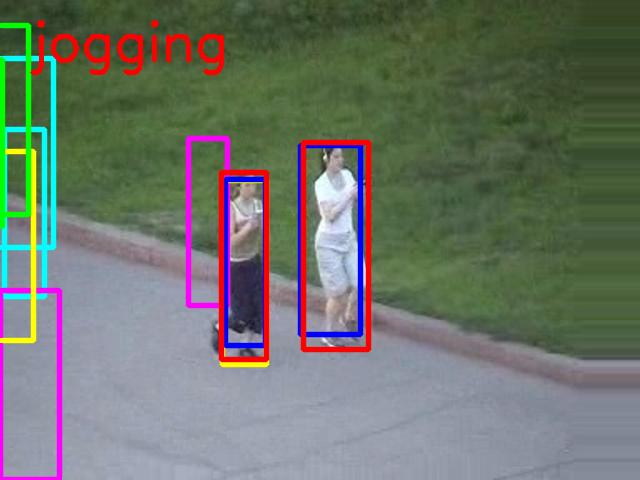}
\includegraphics[width=0.094\linewidth,angle=0]{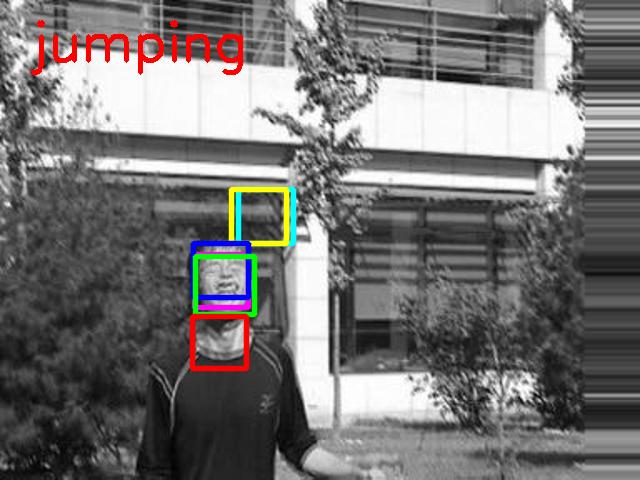}
\includegraphics[width=0.094\linewidth,angle=0]{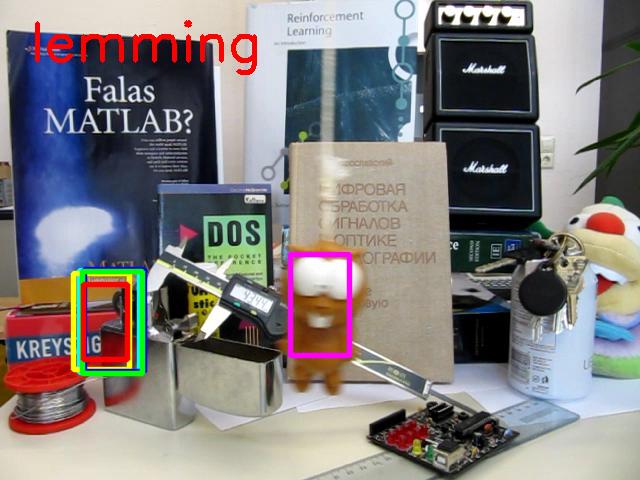}\\
\includegraphics[width=0.094\linewidth,angle=0]{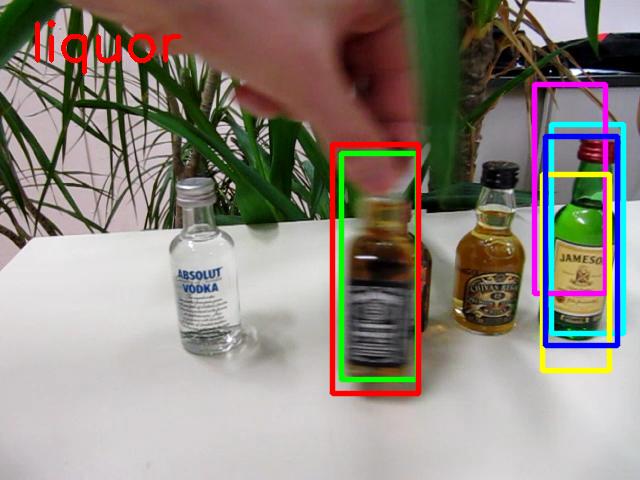}
\includegraphics[width=0.094\linewidth,angle=0]{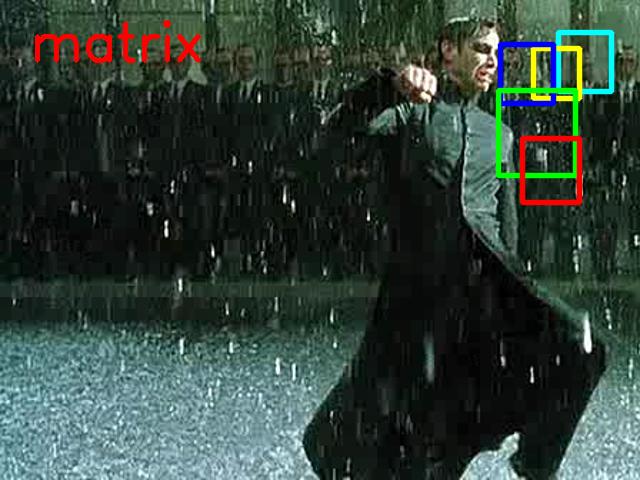}
\includegraphics[width=0.094\linewidth,angle=0]{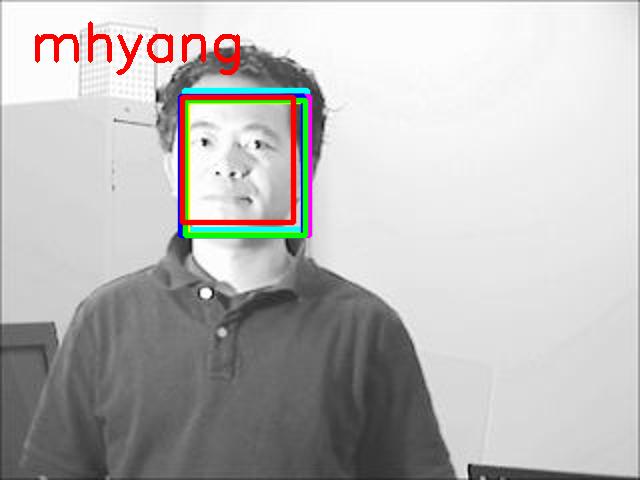}
\includegraphics[width=0.094\linewidth,angle=0]{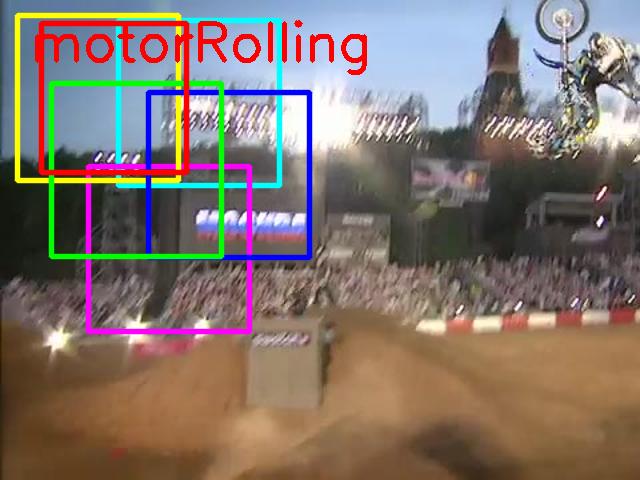}
\includegraphics[width=0.094\linewidth,angle=0]{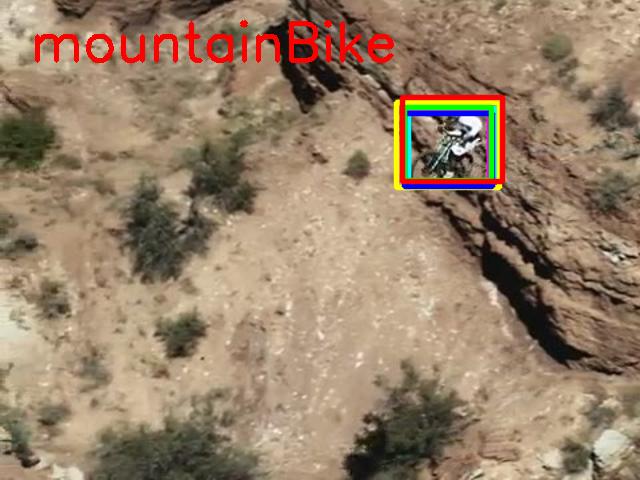}
\includegraphics[width=0.094\linewidth,angle=0]{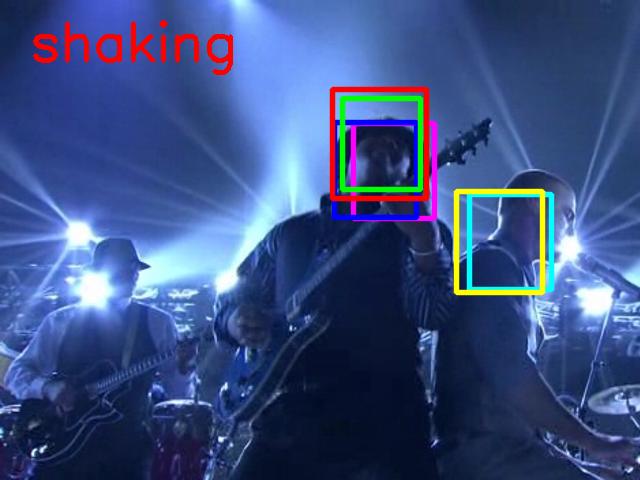}
\includegraphics[width=0.094\linewidth,angle=0]{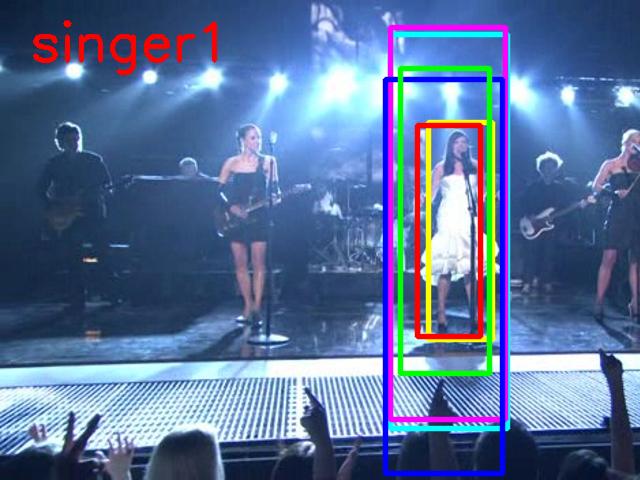}
\includegraphics[width=0.094\linewidth,angle=0]{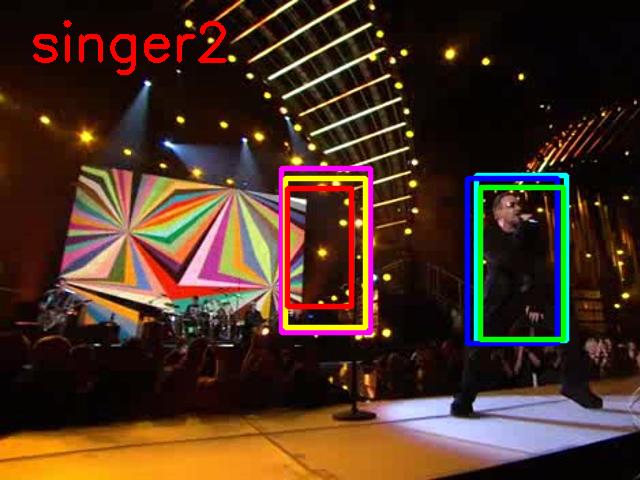}
\includegraphics[width=0.094\linewidth,angle=0]{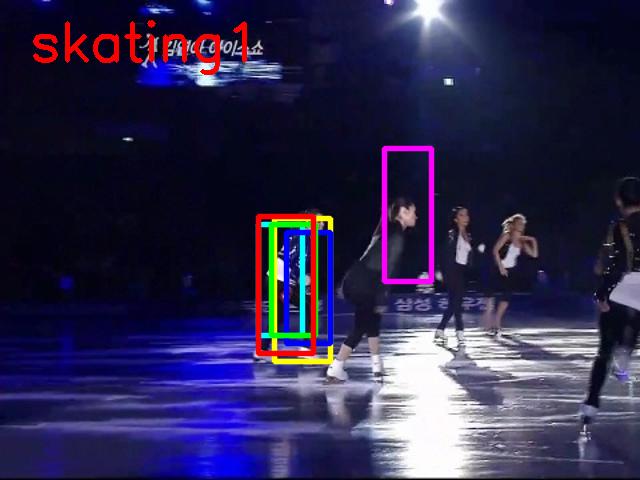}
\includegraphics[width=0.094\linewidth,angle=0]{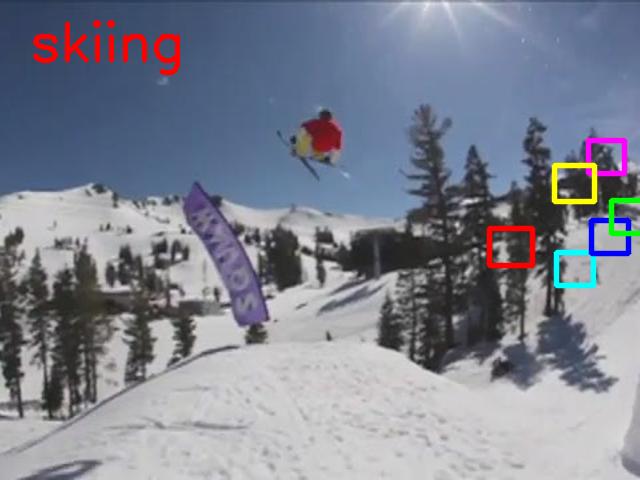}\\
\includegraphics[width=0.094\linewidth,angle=0]{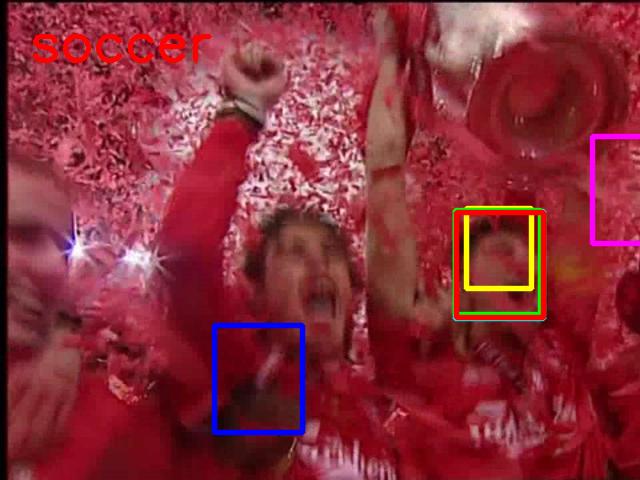}
\includegraphics[width=0.094\linewidth,angle=0]{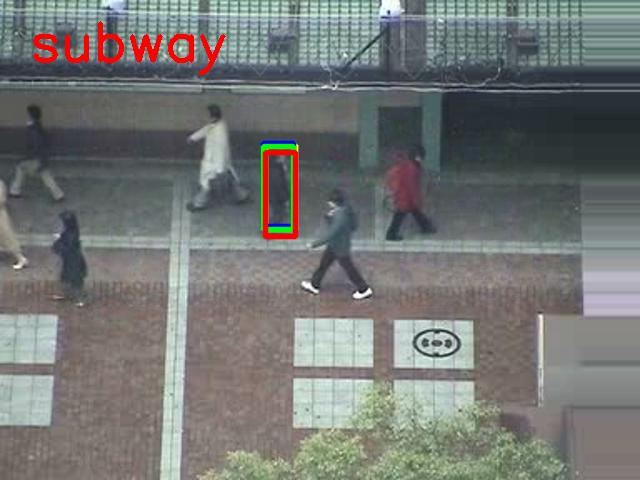}
\includegraphics[width=0.094\linewidth,angle=0]{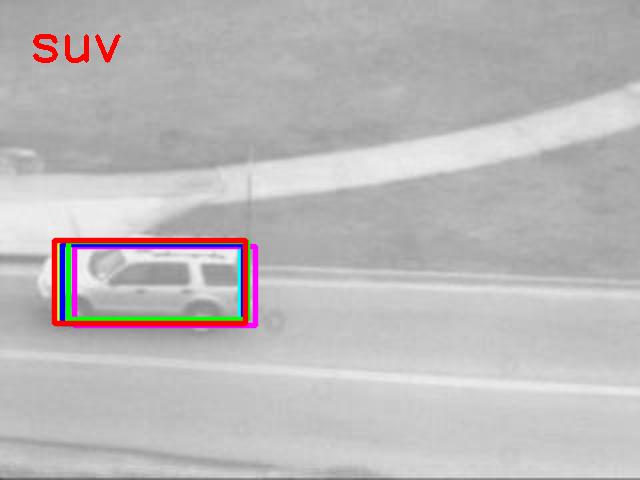}
\includegraphics[width=0.094\linewidth,angle=0]{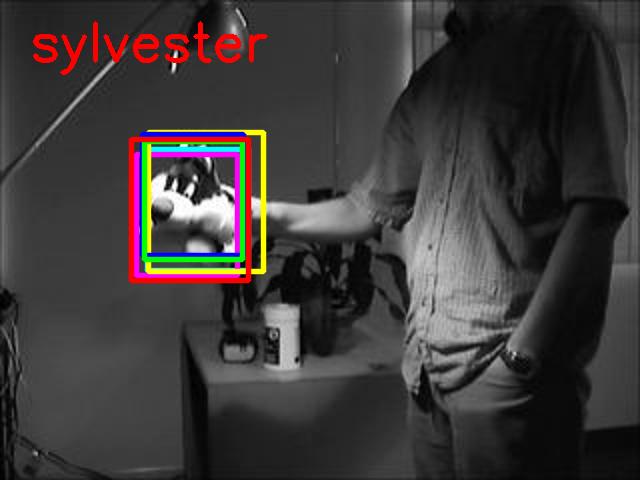}
\includegraphics[width=0.094\linewidth,angle=0]{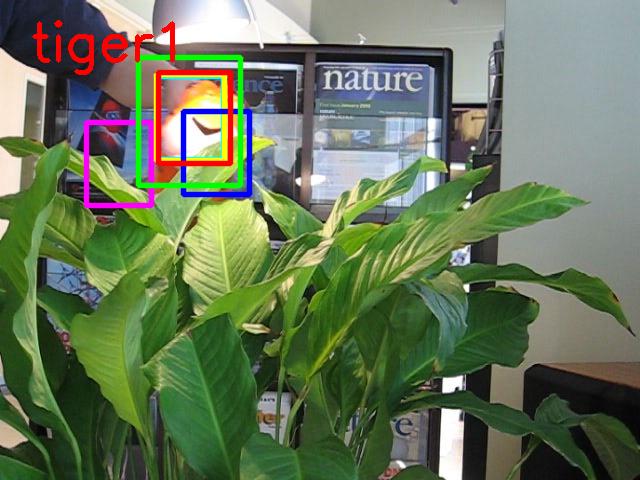}
\includegraphics[width=0.094\linewidth,angle=0]{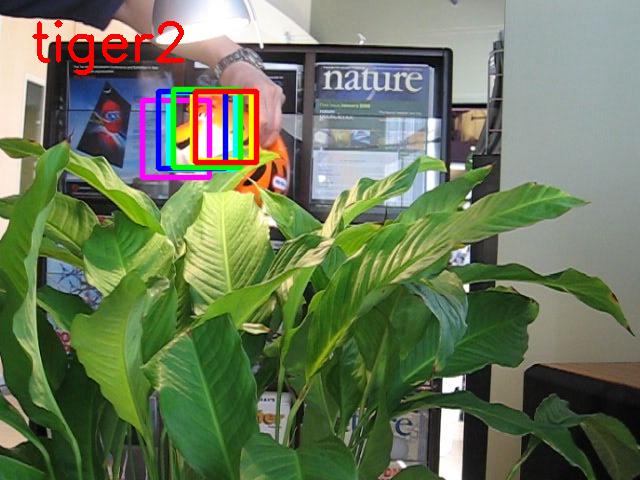}
\includegraphics[width=0.094\linewidth,angle=0]{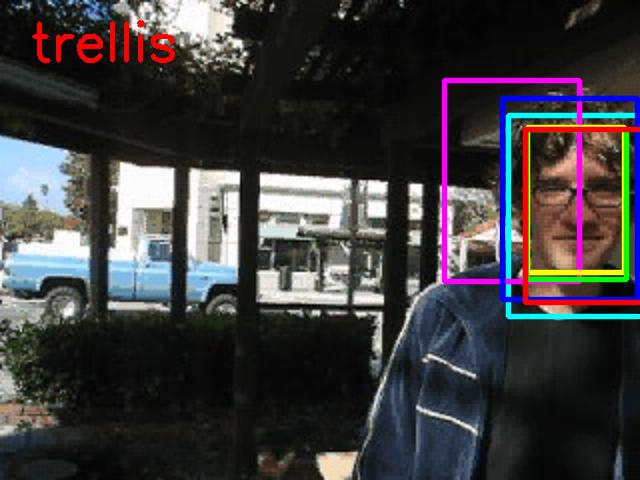}
\includegraphics[width=0.094\linewidth,angle=0]{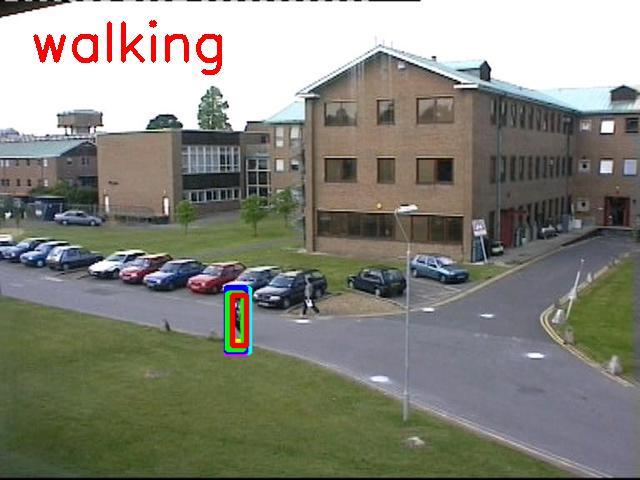}
\includegraphics[width=0.094\linewidth,angle=0]{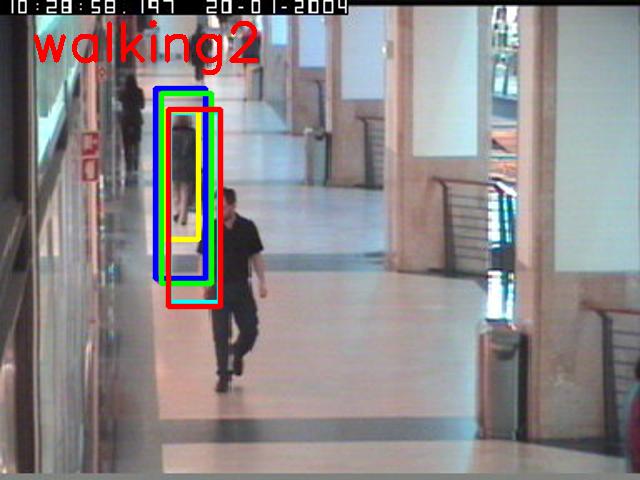}
\includegraphics[width=0.094\linewidth,angle=0]{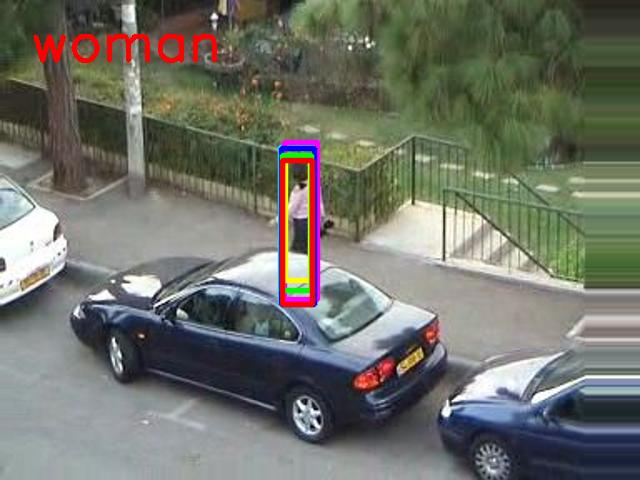}
  \caption{Qualitative evaluations on the CVPR2013 Tracking Benchmark. We compare BIT with the top-performing {\color{green}{RPT}}, {\color{blue}{TGPR}}, {\color{yellow}{ICF}}, {\color[rgb]{1,0,1}{KCF}} and {\color{cyan}{Struck}}.}
  \label{fig:sequences}
\end{figure*}

The CVPR2013 Visual Tracker Benchmark (TB2013) \cite{cvpr2013} contains 50 fully annotated sequences, as shown in Fig. \ref{fig:sequences}. These sequences include many popular sequences used in the online tracking literature over the past several years. For better evaluation and analysis of the strength and weakness of tracking approaches, these sequences are annotated with 11 attributes: illumination variation (IV), scale variation (SV), occlusion (OCC), deformation (DEF), motion blur (MB), fast motion (FM), in-plane rotation (IPR), out-of-plane rotation (OPR), out-of-view (OV), background clutter (BC), and low resolution (LR). In this paper, we compare our method with 11 representative tracking methods. Among the competitors, RPT \cite{rpt}, TGPR \cite{tgpr}, ICF \cite{icf} and KCF \cite{kcf} are the most recent state-of-the-art visual trackers; Struck \cite{struck}, SCM \cite{scm}, TLD \cite{tld}, VTS \cite{vts} are the top four methods as reported in the benchmark; IVT \cite{ivt} and MIL \cite{mil} are classical tracking methods which are used as comparison baselines.

The best way to evaluate trackers is still a debatable subject. Averaged measures like mean center location error or average bounding box overlap penalize an accurate tracker that briefly fails more than they penalize an inaccurate tracker. According to \cite{cvpr2013}, the evaluation for the robustness of trackers is based on two different metrics: the precision plot and success plot. The precision plot shows the percentage of frames on which the Center Location Error (CLE) of a tracker is within a given threshold $e$, where CLE is defined as the center distance between the tracker output and the ground truth. The success plot also counts the percentage of successfully tracked frames by measuring the Intersection Over Union (IOU) metrics on each frame, and the ranking of trackers is based on the Area Under Curve (AUC) score. Following the setting in \cite{cvpr2013}, we conduct the experiment using the one-pass evaluation (OPE) strategy for a better comparison with the latest methods.
\begin{figure}[!t]
\centering
\includegraphics[width=0.49\linewidth]{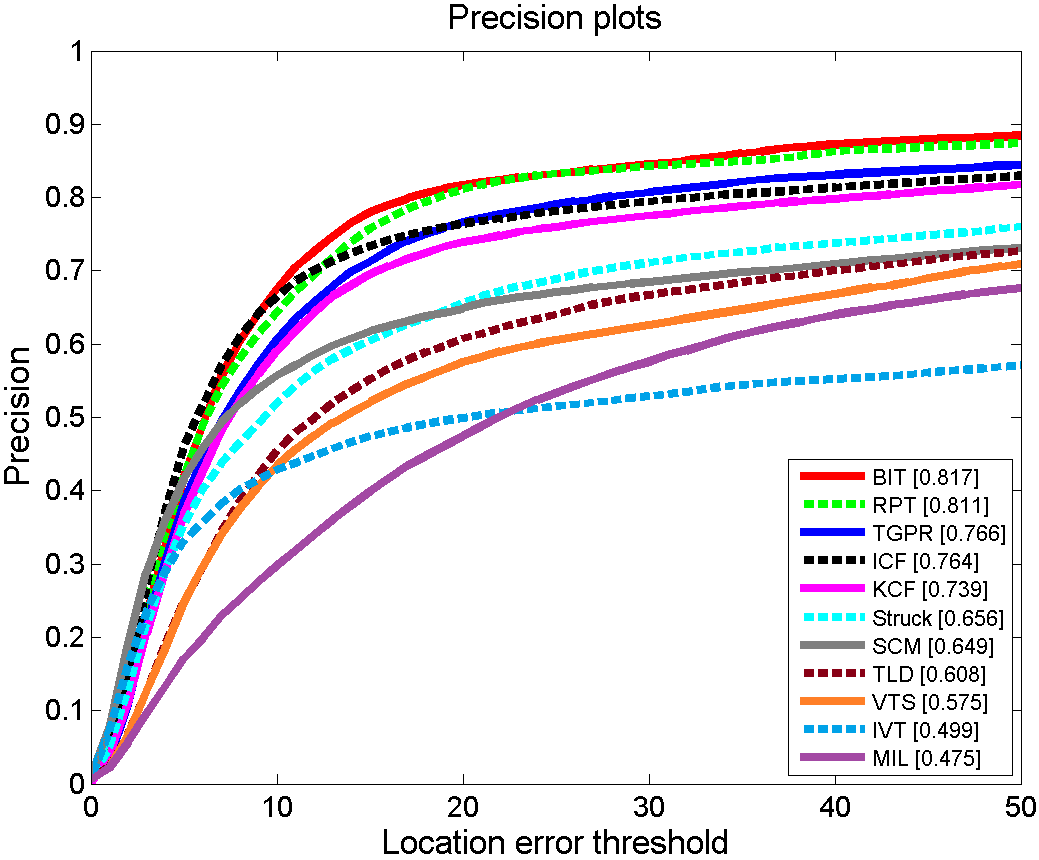}
\includegraphics[width=0.49\linewidth]{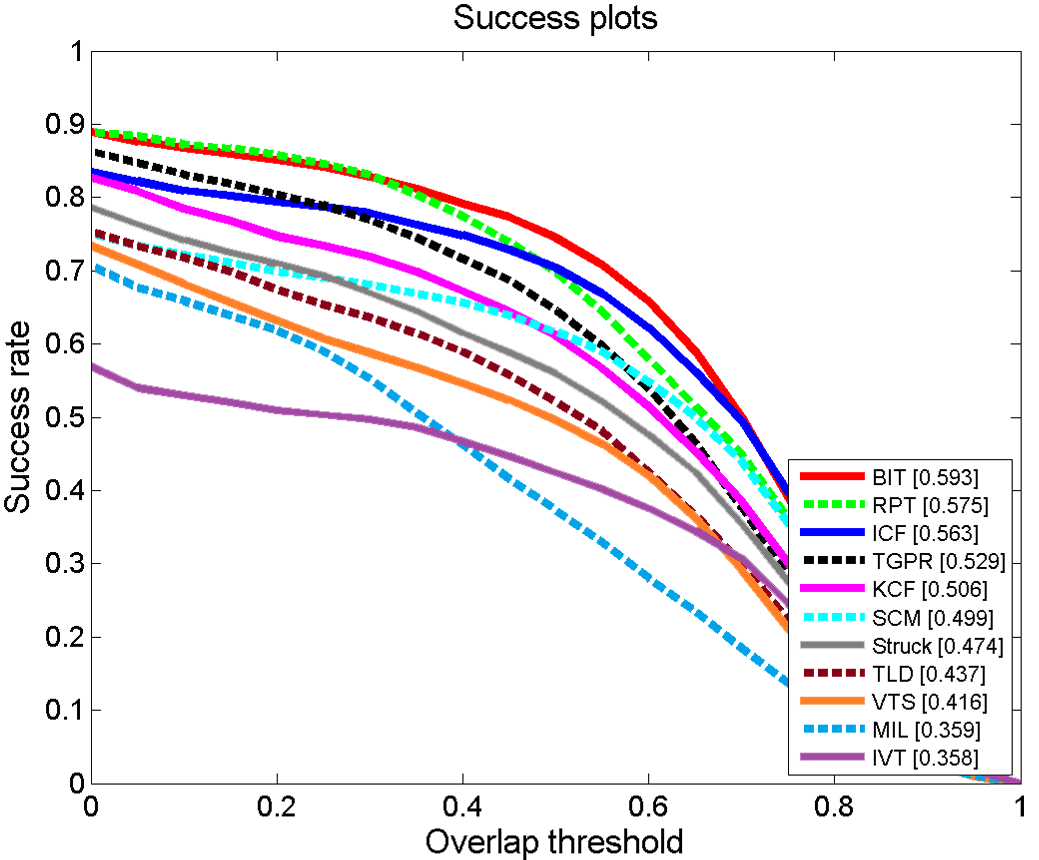}
\caption{Precisions plots and success plots on the TB2013}
\label{fig:ope}
\end{figure}

Fig. \ref{fig:ope} shows the qualitative comparison with selected trackers over all 50 sequences on the TB2013. BIT mainly focuses on the position of the bounding box and ensures the scale robustness by multi-scale filters in S1 units, resulting in the lack of size adjustment of the bounding box. The AUC score is sensitive to bounding box size. In the success plot, a scale estimation method \cite{ddst} is only used to estimate bounding box size, not to aid target location and model updating. However, many trackers (eg. TGPR \cite{tgpr}, KCF \cite{kcf}, Struck \cite{struck}) are the lack of size adjustment of the bounding box, so the precision plot is emphatically analyzed. According to the precision plot ranked by a representative precision score ($e = 20$), our method achieves better average performance than other trackers. The performance gap between our method and the reported best result in the literature is 0.6\% for the tracking precision measure; our method achieves 81.7\% accuracy while the best state-of-the-art is 81.1\% (RPT \cite{rpt}). Moreover the BIT significantly outperforms the best tracker in the benchmark \cite{cvpr2013} by 15.4\% (Struck \cite{struck}) in mean CLE at the threshold of 20 pixels. The results for all the trackers and all the video sequences are given in Table \ref{tab:ope}. Our biologically inspired tracker ranks as the best method 23 times. The equivalent number of best place rankings for RPT, TGPR, ICF and KCF are 23, 11, 20 and 15 respectively. Another observation from Table \ref{tab:ope} is that the BIT rarely performs inaccurately; there are only eight occasions when the proposed tracker performs significantly worse than the best method (no less than 80\% of the highest score for one sequence).

\begin{figure*}[!t]
\centering
\includegraphics[width=0.245\linewidth]{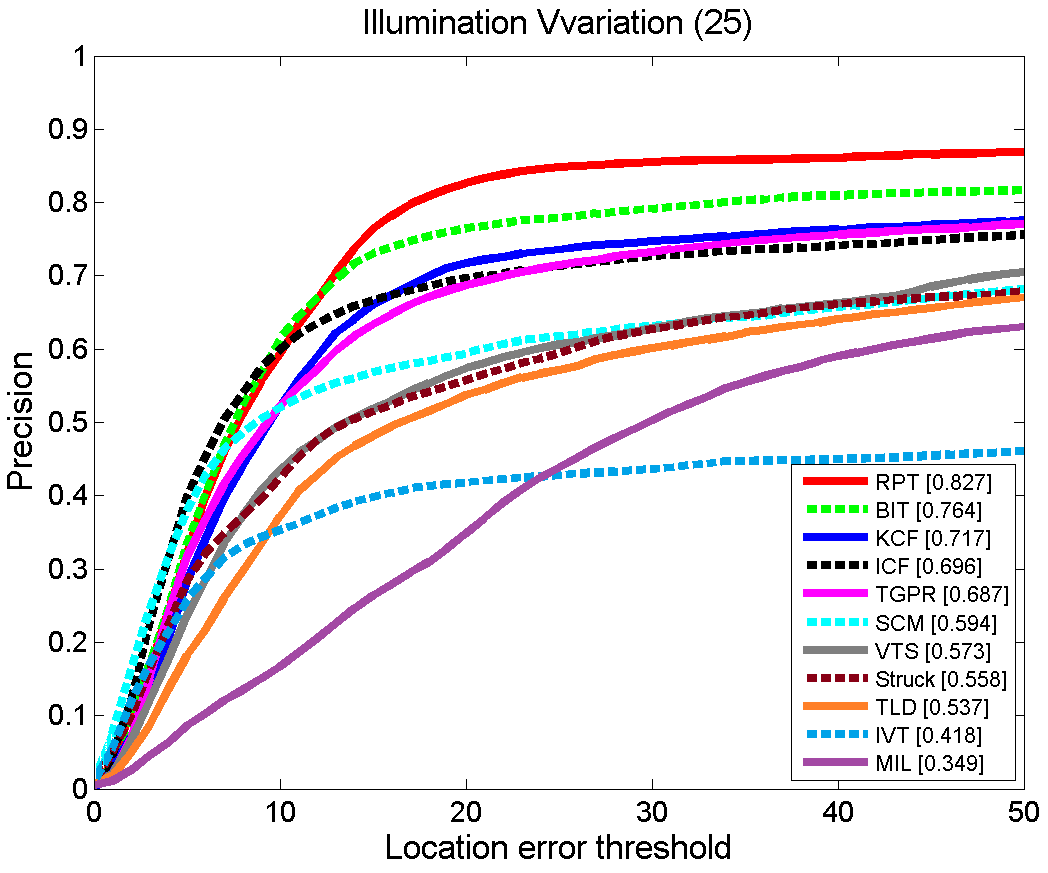}
\includegraphics[width=0.245\linewidth]{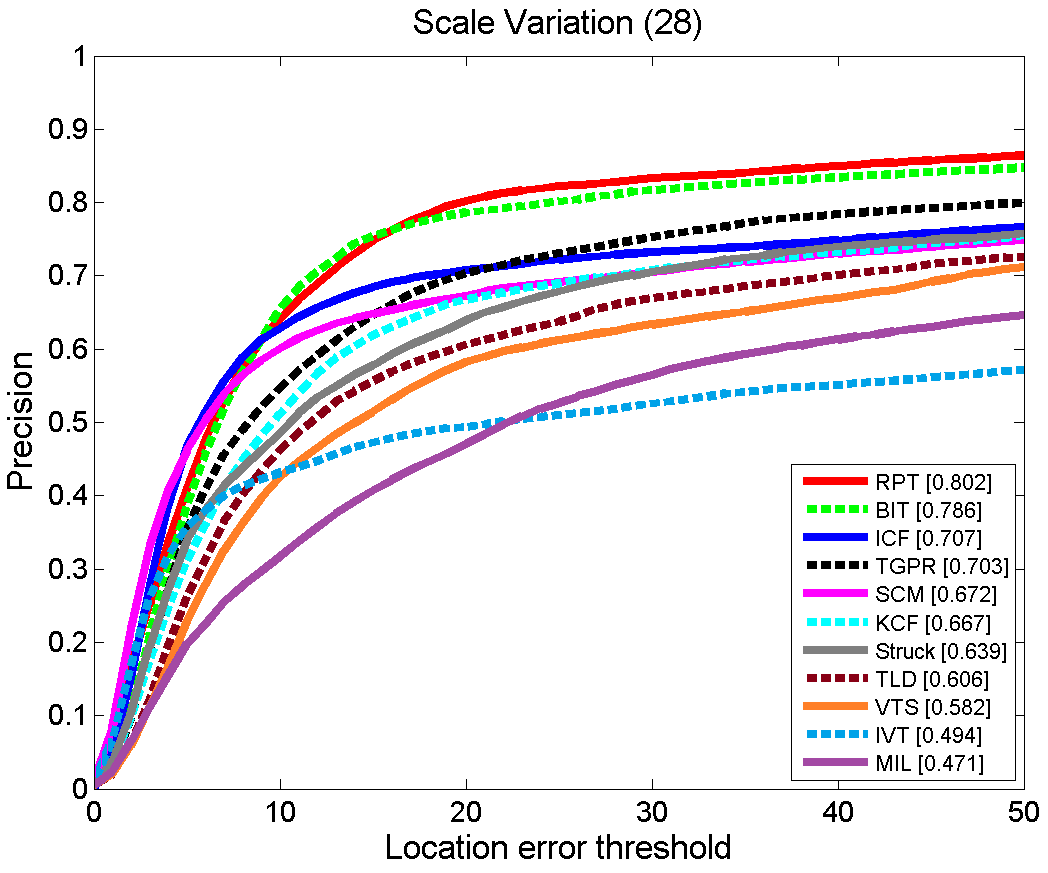}
\includegraphics[width=0.245\linewidth]{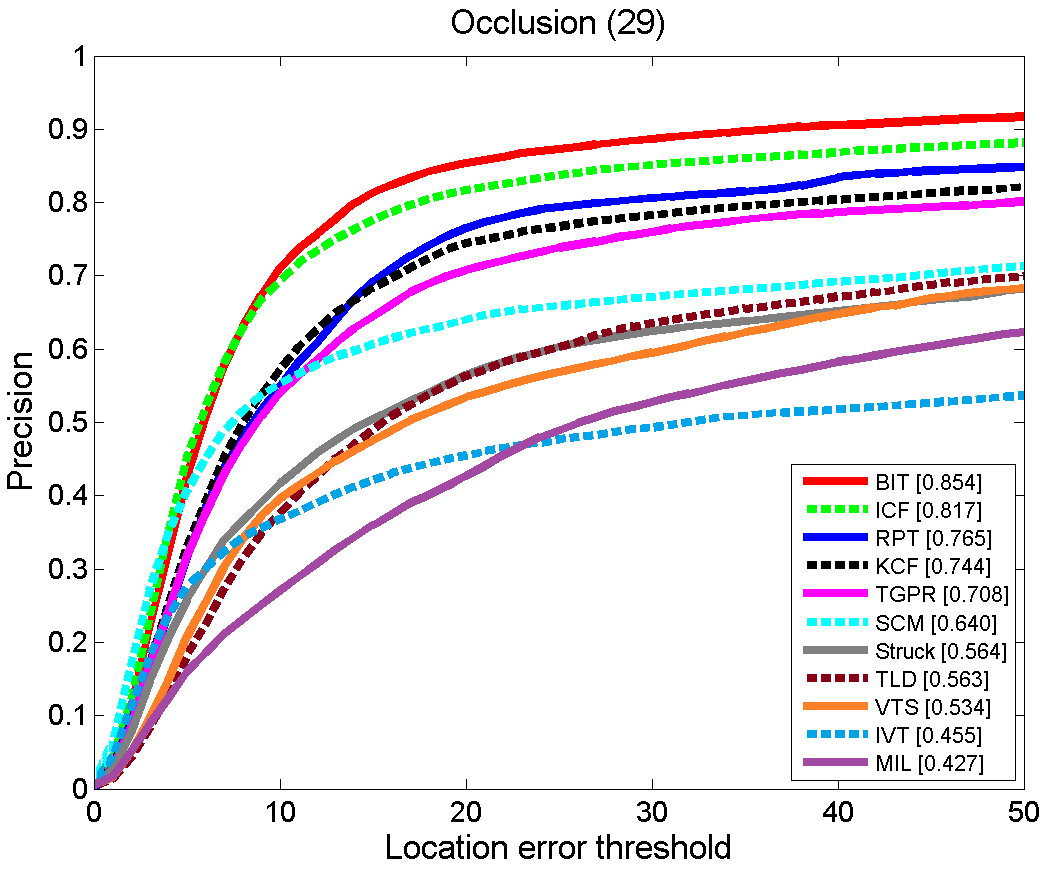}
\includegraphics[width=0.245\linewidth]{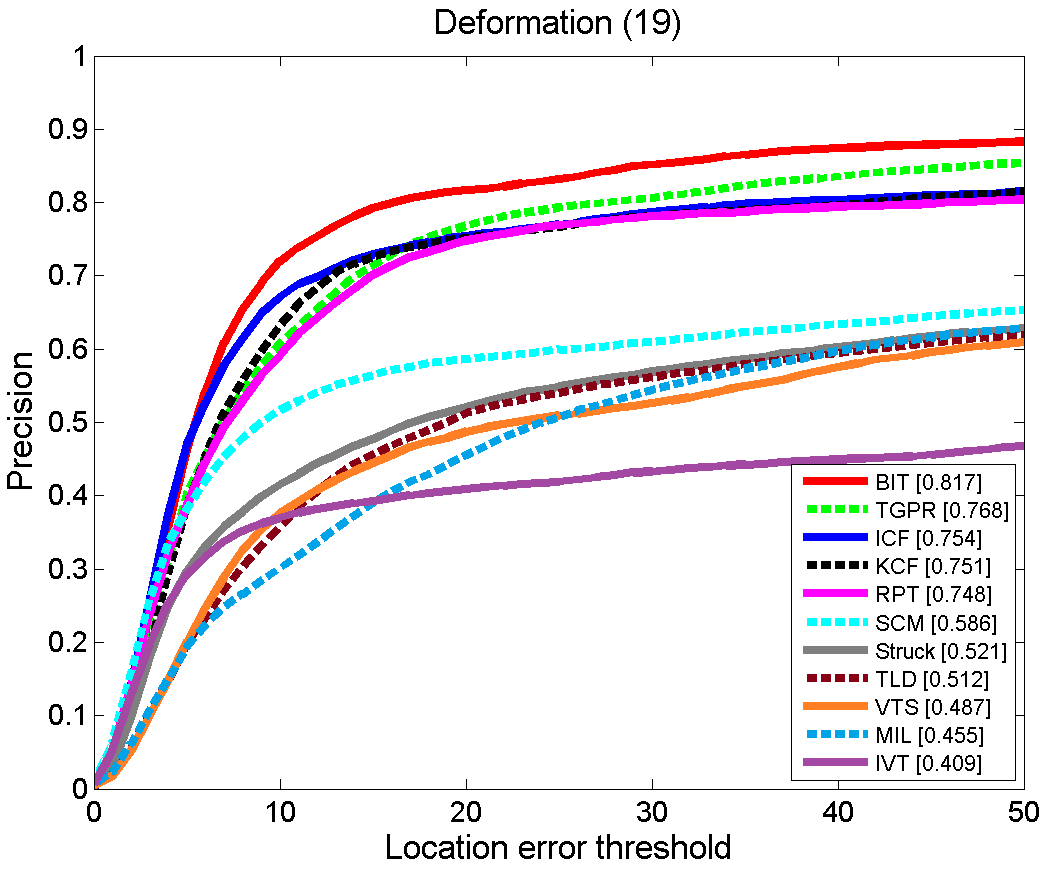}\\
\includegraphics[width=0.245\linewidth]{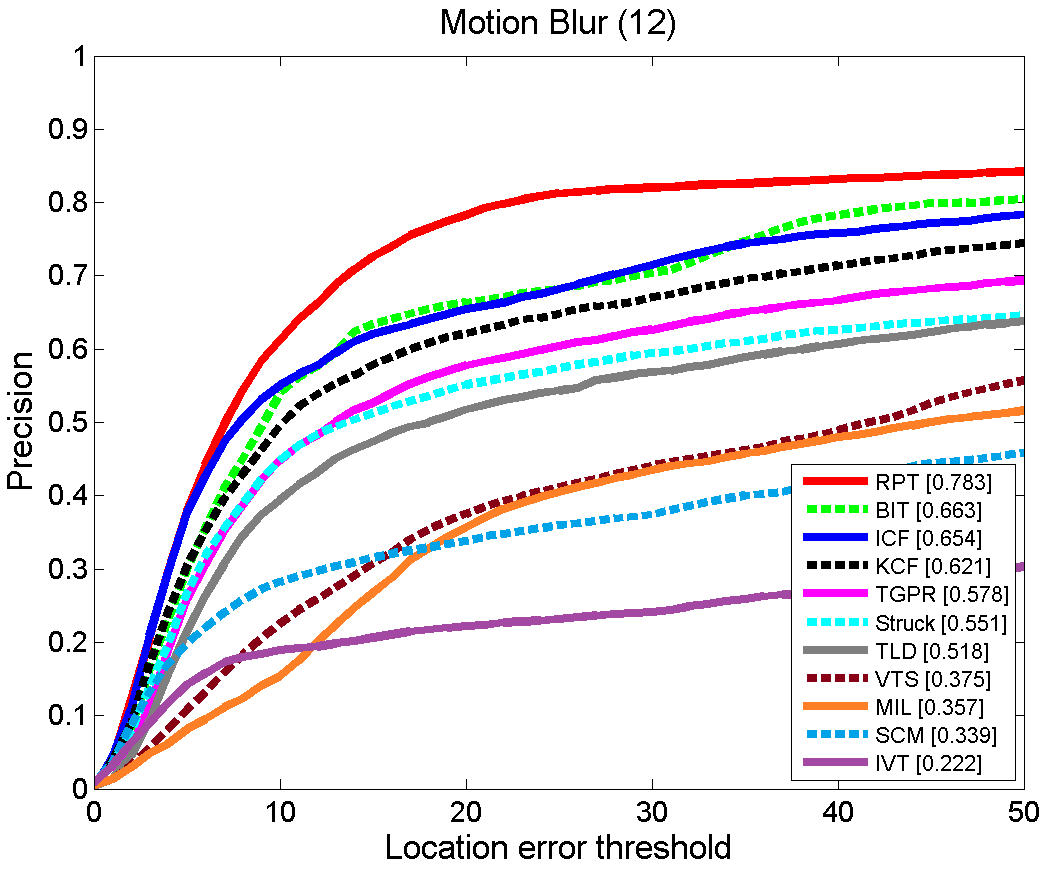}
\includegraphics[width=0.245\linewidth]{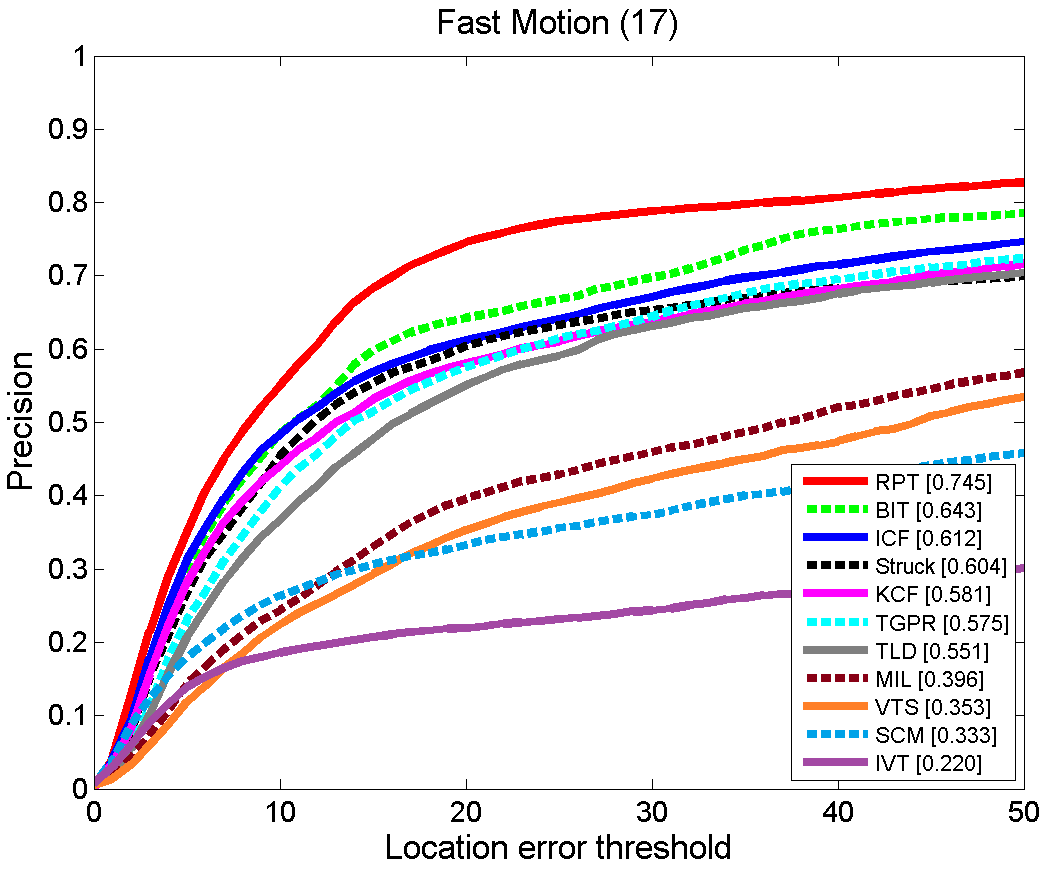}
\includegraphics[width=0.245\linewidth]{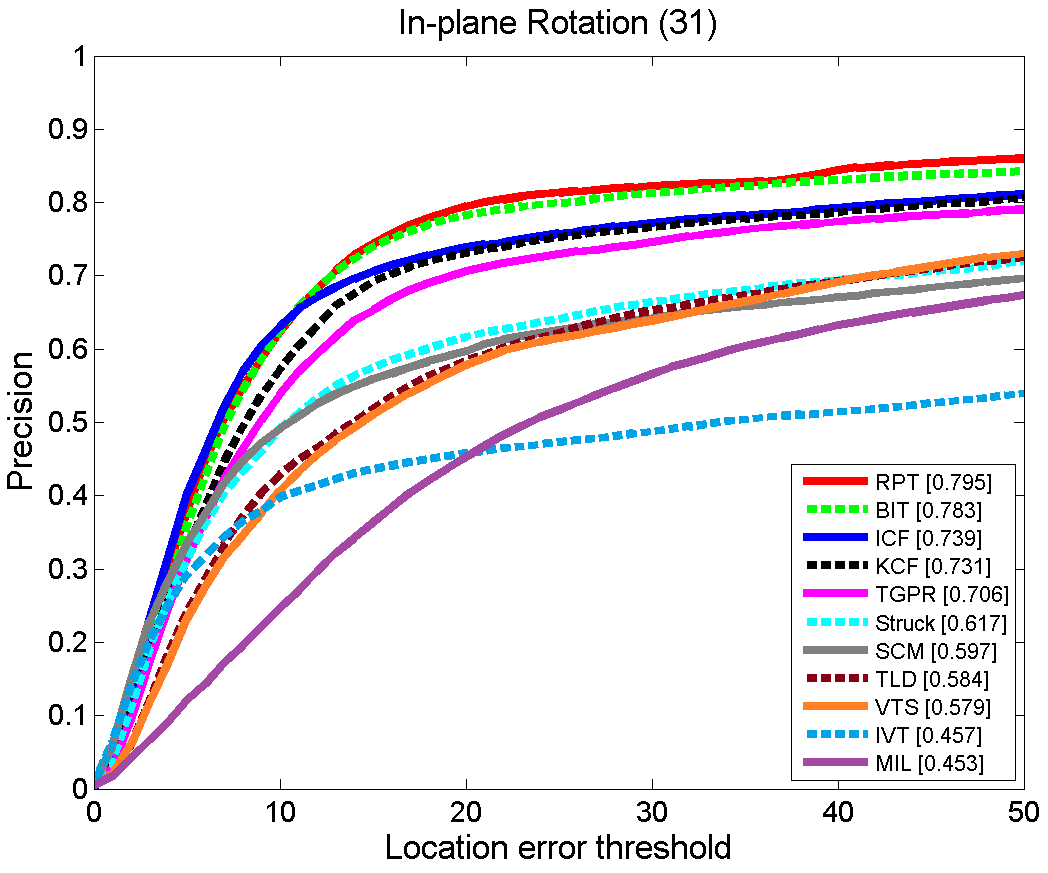}
\includegraphics[width=0.245\linewidth]{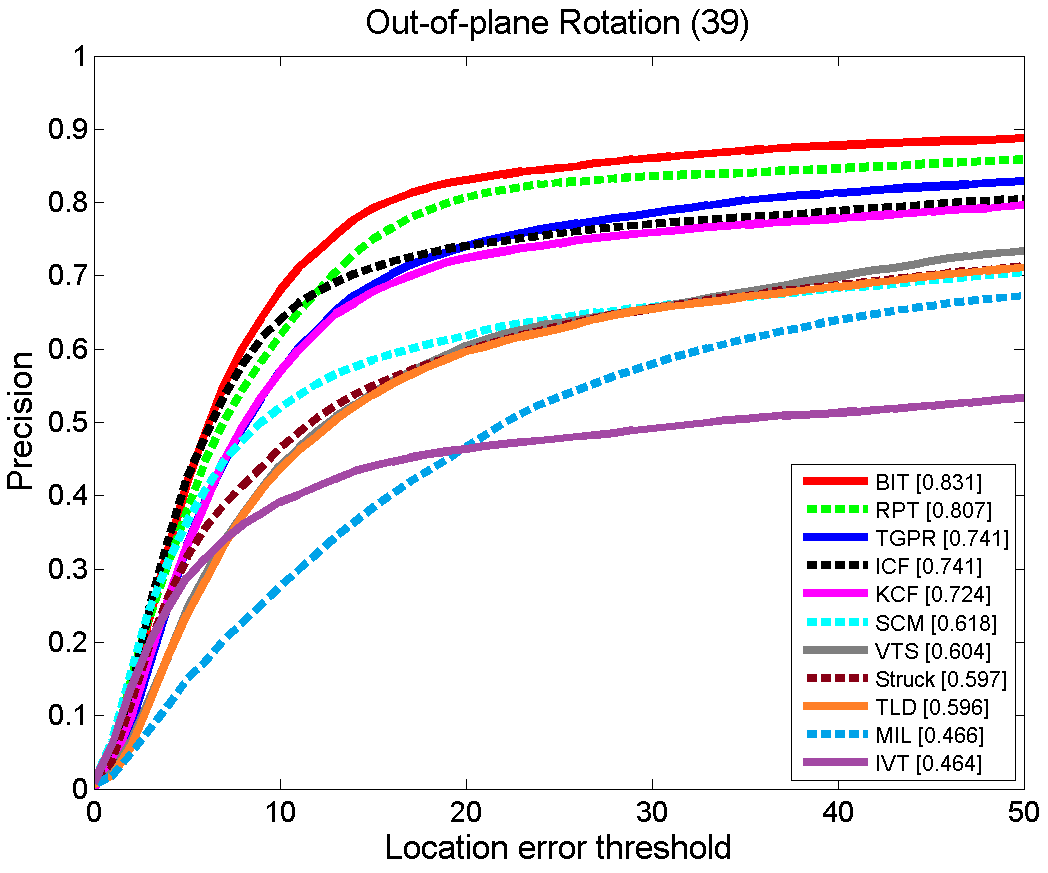}\\
\includegraphics[width=0.245\linewidth]{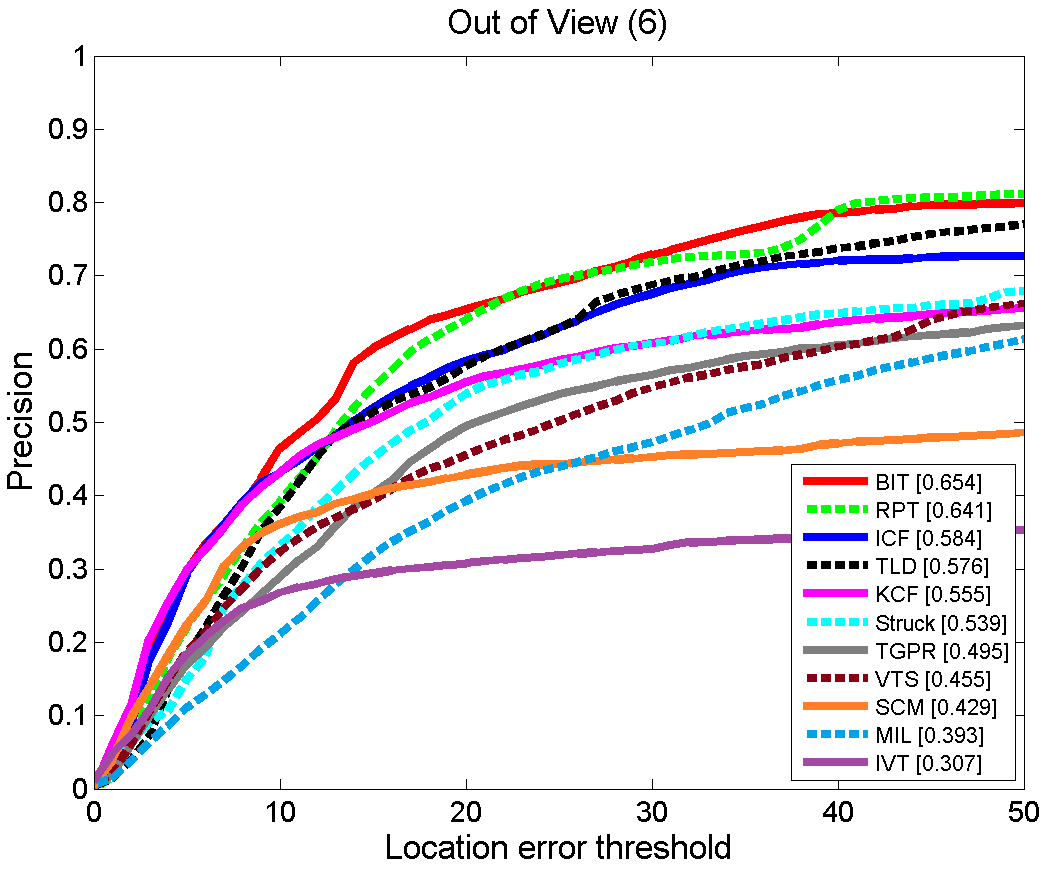}
\includegraphics[width=0.245\linewidth]{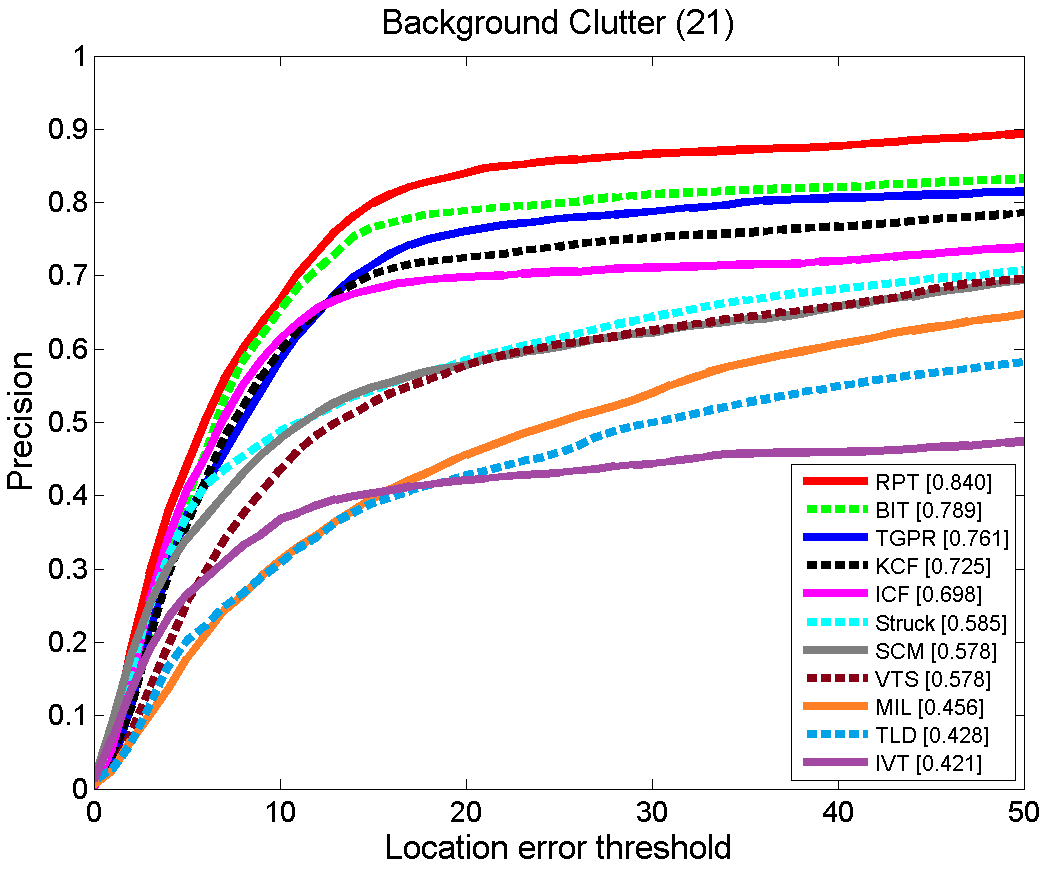}
\includegraphics[width=0.245\linewidth]{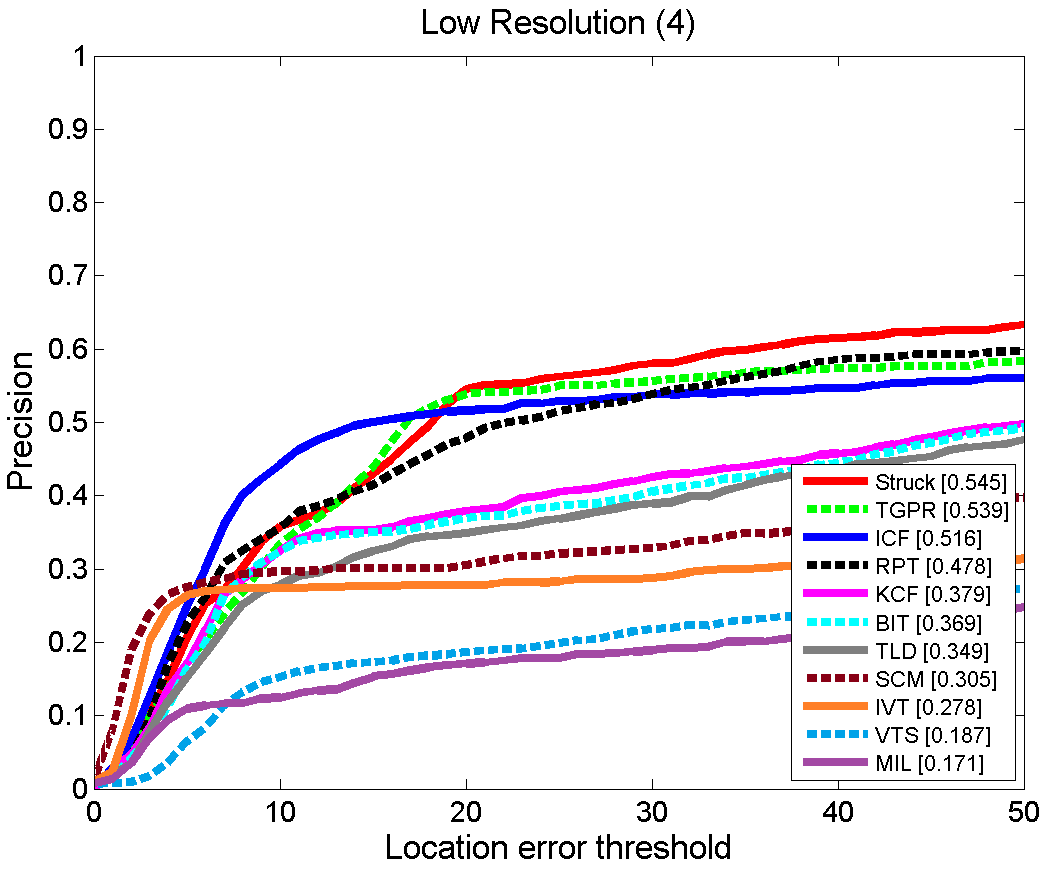}\\
\caption{Precisions plots of the tracking results on the TB2013 for 11 kinds of tracking difficulty}
\label{fig:subsets}
\end{figure*}

Table \ref{tab:subsets} and Fig. \ref{fig:subsets} show the performance plots for 11 kinds of challenge in visual tracking, i.e., fast-motion, background-clutter, motion-blur, deformation, illumination-variation, in-plane-rotation, low-resolution, occlusion, out-of-plane-rotation, out-of-view and scale-variation. Clearly, the BIT almost achieved excellent performances in 11 typical challenge subsets, especially on IV, SV, OCC, DEF, IPR, OPR, and OV. Multi-direction Gabor filters used in S1 units contribute to the robustness of illumination (IV) and rotation (IPR and OPR). Pooling operations in C1 and S2 units provide the shift and scale competitive mechanism to deal with deformation (DEF) and scale variation (SV). Moreover, the generative model in S2 units and the discriminative model in C2 units rise to the challenges of OCC and OV respectively. However, the STD pooling operation in C1 units achieves the shift invariance and at the same time weakens the appearance ability to low resolution (LR) targets.
\begin{table*}
\center
\caption{The tracking scores of the BIT and other visual trackers on the TB2013. The top scores are shown in \textbf{\color{red}{red}} for each row; a score is shown in \underline{\color{blue}{blue}} if it is higher than 80\% of the highest value in that row.}
\begin{tabular}{l|c|c|c|c|c|c|c|c|c|c|c}
\hline
\hline
&BIT&RPT\cite{rpt}&TGPR\cite{tgpr}&ICF\cite{icf}&KCF\cite{kcf}&Struck\cite{struck}&SCM\cite{scm}&TLD\cite{tld}&VTS\cite{vts}&MIL\cite{mil}&IVT\cite{ivt}\\\hline
Basketball&\color{red}{\textbf{1.000}}&\color{blue}{\underline{0.924}}&\color{blue}{\underline{0.994}}&\color{red}{\textbf{1.000}}&\color{blue}{\underline{0.923}}&0.120&0.661&0.028&\color{red}{\textbf{1.000}}&0.284&0.497\\\hline
Bolt&\color{red}{\textbf{1.000}}&0.017&0.017&\color{red}{\textbf{1.000}}&\color{blue}{\underline{0.989}}&0.020&0.031&0.306&0.089&0.014&0.014\\\hline
Boy&\color{red}{\textbf{1.000}}&\color{red}{\textbf{1.000}}&\color{blue}{\underline{0.987}}&\color{red}{\textbf{1.000}}&\color{red}{\textbf{1.000}}&\color{red}{\textbf{1.000}}&0.440&\color{red}{\textbf{1.000}}&\color{blue}{\underline{0.980}}&\color{blue}{\underline{0.846}}&0.332\\\hline
Car4&\color{blue}{\underline{0.973}}&\color{blue}{\underline{0.980}}&\color{red}{\textbf{1.000}}&\color{red}{\textbf{1.000}}&\color{blue}{\underline{0.950}}&\color{blue}{\underline{0.992}}&\color{blue}{\underline{0.974}}&\color{blue}{\underline{0.874}}&0.363&0.354&\color{red}{\textbf{1.000}}\\\hline
CarDark&\color{red}{\textbf{1.000}}&\color{red}{\textbf{1.000}}&\color{red}{\textbf{1.000}}&\color{red}{\textbf{1.000}}&\color{red}{\textbf{1.000}}&\color{red}{\textbf{1.000}}&\color{red}{\textbf{1.000}}&0.639&\color{red}{\textbf{1.000}}&0.379&\color{blue}{\underline{0.807}}\\\hline
CarScale&\color{blue}{\underline{0.718}}&\color{blue}{\underline{0.806}}&\color{blue}{\underline{0.790}}&\color{blue}{\underline{0.806}}&\color{blue}{\underline{0.806}}&0.647&0.647&\color{red}{\textbf{0.853}}&0.544&0.627&\color{blue}{\underline{0.782}}\\\hline
Coke&\color{blue}{\underline{0.931}}&\color{red}{\textbf{0.962}}&\color{blue}{\underline{0.945}}&\color{blue}{\underline{0.887}}&\color{blue}{\underline{0.838}}&\color{blue}{\underline{0.948}}&0.430&0.684&0.189&0.151&0.131\\\hline
Couple&0.607&0.679&0.600&0.107&0.257&0.736&0.114&\color{red}{\textbf{1.000}}&0.100&0.679&0.086\\\hline
Crossing&\color{red}{\textbf{1.000}}&\color{red}{\textbf{1.000}}&\color{red}{\textbf{1.000}}&\color{red}{\textbf{1.000}}&\color{red}{\textbf{1.000}}&\color{red}{\textbf{1.000}}&\color{red}{\textbf{1.000}}&0.617&0.417&\color{red}{\textbf{1.000}}&\color{red}{\textbf{1.000}}\\\hline
David&\color{red}{\textbf{1.000}}&\color{red}{\textbf{1.000}}&\color{blue}{\underline{0.977}}&\color{red}{\textbf{1.000}}&\color{red}{\textbf{1.000}}&0.329&\color{red}{\textbf{1.000}}&\color{red}{\textbf{1.000}}&\color{blue}{\underline{0.962}}&0.699&\color{red}{\textbf{1.000}}\\\hline
David2&\color{red}{\textbf{1.000}}&\color{red}{\textbf{1.000}}&\color{red}{\textbf{1.000}}&\color{red}{\textbf{1.000}}&\color{red}{\textbf{1.000}}&\color{red}{\textbf{1.000}}&\color{red}{\textbf{1.000}}&\color{red}{\textbf{1.000}}&\color{red}{\textbf{1.000}}&\color{blue}{\underline{0.978}}&\color{red}{\textbf{1.000}}\\\hline
David3&\color{red}{\textbf{1.000}}&\color{red}{\textbf{1.000}}&\color{blue}{\underline{0.996}}&\color{red}{\textbf{1.000}}&\color{red}{\textbf{1.000}}&0.337&0.496&0.111&0.742&0.738&0.754\\\hline
Deer&\color{blue}{\underline{0.831}}&\color{red}{\textbf{1.000}}&\color{blue}{\underline{0.859}}&\color{blue}{\underline{0.817}}&\color{blue}{\underline{0.817}}&\color{red}{\textbf{1.000}}&0.028&0.732&0.042&0.127&0.028\\\hline
Dog1&\color{red}{\textbf{1.000}}&\color{red}{\textbf{1.000}}&\color{red}{\textbf{1.000}}&\color{blue}{\underline{0.994}}&\color{red}{\textbf{1.000}}&\color{blue}{\underline{0.996}}&\color{blue}{\underline{0.976}}&\color{red}{\textbf{1.000}}&\color{blue}{\underline{0.811}}&\color{blue}{\underline{0.919}}&\color{blue}{\underline{0.980}}\\\hline
Doll&\color{blue}{\underline{0.986}}&\color{red}{\textbf{0.987}}&\color{blue}{\underline{0.943}}&\color{blue}{\underline{0.947}}&\color{blue}{\underline{0.967}}&\color{blue}{\underline{0.919}}&\color{blue}{\underline{0.978}}&\color{blue}{\underline{0.983}}&\color{blue}{\underline{0.946}}&0.732&0.757\\\hline
Dudek&\color{blue}{\underline{0.862}}&\color{blue}{\underline{0.849}}&\color{blue}{\underline{0.751}}&\color{red}{\textbf{0.899}}&\color{blue}{\underline{0.859}}&\color{blue}{\underline{0.897}}&\color{blue}{\underline{0.883}}&0.597&\color{blue}{\underline{0.871}}&0.688&\color{blue}{\underline{0.886}}\\\hline
FaceOcc1&\color{blue}{\underline{0.877}}&0.663&0.664&\color{blue}{\underline{0.855}}&\color{blue}{\underline{0.878}}&0.575&\color{red}{\textbf{0.933}}&0.203&0.485&0.221&0.645\\\hline
FaceOcc2&\color{blue}{\underline{0.933}}&\color{blue}{\underline{0.990}}&0.468&\color{blue}{\underline{0.968}}&\color{blue}{\underline{0.972}}&\color{red}{\textbf{1.000}}&\color{blue}{\underline{0.860}}&\color{blue}{\underline{0.856}}&\color{blue}{\underline{0.936}}&0.740&\color{blue}{\underline{0.993}}\\\hline
Fish&\color{red}{\textbf{1.000}}&\color{red}{\textbf{1.000}}&\color{blue}{\underline{0.975}}&\color{red}{\textbf{1.000}}&\color{red}{\textbf{1.000}}&\color{red}{\textbf{1.000}}&\color{blue}{\underline{0.863}}&\color{red}{\textbf{1.000}}&\color{blue}{\underline{0.992}}&0.387&\color{red}{\textbf{1.000}}\\\hline
FleetFace&\color{blue}{\underline{0.581}}&\color{blue}{\underline{0.562}}&0.453&\color{blue}{\underline{0.627}}&\color{blue}{\underline{0.556}}&\color{blue}{\underline{0.639}}&\color{blue}{\underline{0.529}}&0.506&\color{red}{\textbf{0.642}}&0.358&0.264\\\hline
Football&\color{blue}{\underline{0.798}}&\color{blue}{\underline{0.801}}&\color{red}{\textbf{0.997}}&\color{blue}{\underline{0.801}}&0.796&0.751&0.765&\color{blue}{\underline{0.804}}&0.796&0.790&0.793\\\hline
Football1&\color{blue}{\underline{0.973}}&\color{blue}{\underline{0.932}}&\color{blue}{\underline{0.986}}&\color{blue}{\underline{0.986}}&\color{blue}{\underline{0.959}}&\color{red}{\textbf{1.000}}&0.568&0.554&\color{blue}{\underline{0.892}}&\color{red}{\textbf{1.000}}&\color{blue}{\underline{0.811}}\\\hline
Freeman1&\color{red}{\textbf{1.000}}&\color{blue}{\underline{0.972}}&\color{blue}{\underline{0.933}}&0.393&0.393&\color{blue}{\underline{0.801}}&\color{blue}{\underline{0.982}}&0.540&\color{blue}{\underline{0.969}}&\color{blue}{\underline{0.939}}&\color{blue}{\underline{0.807}}\\\hline
Freeman3&\color{blue}{\underline{0.817}}&\color{blue}{\underline{0.996}}&0.774&\color{blue}{\underline{0.896}}&\color{blue}{\underline{0.911}}&0.789&\color{red}{\textbf{1.000}}&0.767&0.702&0.048&0.761\\\hline
Freeman4&\color{red}{\textbf{0.993}}&\color{blue}{\underline{0.880}}&0.580&\color{blue}{\underline{0.951}}&0.530&0.375&0.509&0.410&0.219&0.201&0.346\\\hline
Girl&\color{red}{\textbf{1.000}}&\color{blue}{\underline{0.924}}&\color{blue}{\underline{0.918}}&\color{blue}{\underline{0.916}}&\color{blue}{\underline{0.864}}&\color{red}{\textbf{1.000}}&\color{red}{\textbf{1.000}}&\color{blue}{\underline{0.918}}&\color{blue}{\underline{0.874}}&0.714&0.444\\\hline
Ironman&0.157&0.181&\color{blue}{\underline{0.217}}&\color{blue}{\underline{0.199}}&\color{blue}{\underline{0.217}}&0.114&0.157&0.120&\color{red}{\textbf{0.247}}&0.108&0.054\\\hline
Jogging.1&\color{blue}{\underline{0.977}}&0.228&\color{red}{\textbf{0.993}}&\color{blue}{\underline{0.977}}&0.235&0.241&0.228&\color{blue}{\underline{0.974}}&0.225&0.231&0.225\\\hline
Jogging.2&\color{red}{\textbf{1.000}}&0.179&\color{blue}{\underline{0.997}}&0.186&0.163&0.254&\color{red}{\textbf{1.000}}&\color{blue}{\underline{0.857}}&0.186&0.186&0.199\\\hline
Jumping&0.093&\color{red}{\textbf{1.000}}&\color{blue}{\underline{0.946}}&0.383&0.339&\color{red}{\textbf{1.000}}&0.153&\color{red}{\textbf{1.000}}&0.236&\color{blue}{\underline{0.997}}&0.208\\\hline
Lemming&0.491&0.537&0.349&0.509&0.495&0.628&0.166&\color{red}{\textbf{0.859}}&0.554&\color{blue}{\underline{0.823}}&0.167\\\hline
Liquor&\color{red}{\textbf{0.986}}&\color{blue}{\underline{0.937}}&0.271&0.431&0.423&0.390&0.276&0.588&0.364&0.199&0.207\\\hline
Matrix&\color{blue}{\underline{0.360}}&\color{red}{\textbf{0.440}}&\color{blue}{\underline{0.390}}&0.350&0.170&0.120&0.350&0.160&0.200&0.180&0.020\\\hline
Mhyang&\color{red}{\textbf{1.000}}&\color{red}{\textbf{1.000}}&\color{blue}{\underline{0.947}}&\color{red}{\textbf{1.000}}&\color{red}{\textbf{1.000}}&\color{red}{\textbf{1.000}}&\color{red}{\textbf{1.000}}&\color{blue}{\underline{0.978}}&\color{red}{\textbf{1.000}}&0.460&\color{red}{\textbf{1.000}}\\\hline
MotorRolling&0.049&0.049&0.091&0.049&0.043&0.085&0.037&\color{red}{\textbf{0.116}}&0.049&0.043&0.030\\\hline
MountainBike&\color{blue}{\underline{0.987}}&\color{red}{\textbf{1.000}}&\color{red}{\textbf{1.000}}&\color{red}{\textbf{1.000}}&\color{red}{\textbf{1.000}}&\color{blue}{\underline{0.921}}&\color{blue}{\underline{0.969}}&0.259&\color{blue}{\underline{0.996}}&0.667&\color{blue}{\underline{0.996}}\\\hline
Shaking&\color{blue}{\underline{0.970}}&\color{red}{\textbf{0.995}}&\color{blue}{\underline{0.970}}&0.025&0.025&0.192&\color{blue}{\underline{0.814}}&0.405&\color{blue}{\underline{0.921}}&0.282&0.011\\\hline
Singer1&\color{red}{\textbf{1.000}}&\color{blue}{\underline{0.986}}&0.684&0.689&\color{blue}{\underline{0.980}}&0.641&\color{red}{\textbf{1.000}}&\color{red}{\textbf{1.000}}&\color{red}{\textbf{1.000}}&0.501&\color{blue}{\underline{0.963}}\\\hline
Singer2&0.036&\color{blue}{\underline{0.913}}&\color{red}{\textbf{0.970}}&0.038&\color{blue}{\underline{0.945}}&0.036&0.112&0.071&0.358&0.404&0.036\\\hline
Skating1&\color{red}{\textbf{1.000}}&\color{red}{\textbf{1.000}}&\color{blue}{\underline{0.805}}&\color{red}{\textbf{1.000}}&\color{red}{\textbf{1.000}}&0.465&0.768&0.318&\color{blue}{\underline{0.890}}&0.130&0.108\\\hline
Skiing&\color{red}{\textbf{0.136}}&\color{red}{\textbf{0.136}}&\color{blue}{\underline{0.123}}&\color{blue}{\underline{0.111}}&0.074&0.037&\color{red}{\textbf{0.136}}&\color{blue}{\underline{0.123}}&0.062&0.074&\color{blue}{\underline{0.111}}\\\hline
Soccer&\color{blue}{\underline{0.949}}&\color{blue}{\underline{0.944}}&0.158&\color{red}{\textbf{0.967}}&\color{blue}{\underline{0.793}}&0.253&0.268&0.115&0.505&0.191&0.173\\\hline
Subway&\color{red}{\textbf{1.000}}&\color{red}{\textbf{1.000}}&\color{red}{\textbf{1.000}}&\color{red}{\textbf{1.000}}&\color{red}{\textbf{1.000}}&\color{blue}{\underline{0.983}}&\color{red}{\textbf{1.000}}&0.251&0.240&\color{blue}{\underline{0.994}}&0.223\\\hline
Suv&\color{red}{\textbf{0.979}}&0.529&0.658&\color{red}{\textbf{0.979}}&\color{red}{\textbf{0.979}}&0.572&\color{blue}{\underline{0.978}}&\color{blue}{\underline{0.909}}&0.535&0.123&0.447\\\hline
Sylvester&\color{blue}{\underline{0.839}}&\color{blue}{\underline{0.979}}&\color{blue}{\underline{0.955}}&\color{blue}{\underline{0.851}}&\color{blue}{\underline{0.843}}&\color{red}{\textbf{0.995}}&\color{blue}{\underline{0.946}}&\color{blue}{\underline{0.949}}&\color{blue}{\underline{0.820}}&0.651&0.680\\\hline
Tiger1&\color{blue}{\underline{0.927}}&\color{red}{\textbf{0.977}}&0.284&\color{blue}{\underline{0.958}}&\color{blue}{\underline{0.975}}&0.175&0.126&0.456&0.117&0.095&0.080\\\hline
Tiger2&0.449&\color{red}{\textbf{0.814}}&\color{blue}{\underline{0.723}}&0.485&0.356&0.630&0.112&0.386&0.162&0.414&0.082\\\hline
Trellis&\color{red}{\textbf{1.000}}&\color{red}{\textbf{1.000}}&\color{blue}{\underline{0.979}}&\color{red}{\textbf{1.000}}&\color{red}{\textbf{1.000}}&\color{blue}{\underline{0.877}}&\color{blue}{\underline{0.873}}&0.529&0.503&0.230&0.332\\\hline
Walking&\color{red}{\textbf{1.000}}&\color{red}{\textbf{1.000}}&\color{red}{\textbf{1.000}}&\color{red}{\textbf{1.000}}&\color{red}{\textbf{1.000}}&\color{red}{\textbf{1.000}}&\color{red}{\textbf{1.000}}&\color{blue}{\underline{0.964}}&\color{red}{\textbf{1.000}}&\color{red}{\textbf{1.000}}&\color{red}{\textbf{1.000}}\\\hline
Walking2&0.440&0.684&\color{blue}{\underline{0.988}}&\color{red}{\textbf{1.000}}&0.440&\color{blue}{\underline{0.982}}&\color{red}{\textbf{1.000}}&0.426&0.408&0.406&\color{red}{\textbf{1.000}}\\\hline
Woman&\color{blue}{\underline{0.940}}&\color{blue}{\underline{0.938}}&\color{blue}{\underline{0.968}}&\color{blue}{\underline{0.938}}&\color{blue}{\underline{0.938}}&\color{red}{\textbf{1.000}}&\color{blue}{\underline{0.940}}&0.191&0.198&0.206&0.201\\\hline\hline
\textbf{ALL}&\textbf{0.817}&0.811&0.766&0.764&0.739&0.656&0.649&0.608&0.575&0.475&0.499\\\hline
\textbf{No. Best}&\textbf{23}&\textbf{23}&11&20&15&14&14&11&8&3&8\\\hline
\textbf{No. Worst}&\textbf{8}&10&14&12&15&26&23&28&29&41&33\\\hline\hline
\end{tabular}
\label{tab:ope}
\end{table*}
\begin{table*}
\center
\caption{The tracking scores on the TB2013 for 11 kinds of tracking difficulty. \textbf{\color{red}{Red}} indicates the best while \underline{\color{blue}{blue}} indicates the second best.}
\begin{tabular}{l|c|c|c|c|c|c|c|c|c|c|c}
\hline
\hline
&BIT&RPT\cite{rpt}&TGPR\cite{tgpr}&ICF\cite{icf}&KCF\cite{kcf}&Struck\cite{struck}&SCM\cite{scm}&TLD\cite{tld}&VTS\cite{vts}&MIL\cite{mil}&IVT\cite{ivt}\\\hline
IV&\color{blue}{\underline{0.764}}&\color{red}{\textbf{0.827}}&0.687&0.696&0.717&0.558&0.594&0.537&0.573&0.349&0.418\\\hline
SV&\color{blue}{\underline{0.786}}&\color{red}{\textbf{0.802}}&0.703&0.707&0.667&0.639&0.672&0.606&0.582&0.471&0.494\\\hline
OCC&\color{red}{\textbf{0.854}}&0.765&0.708&\color{blue}{\underline{0.817}}&0.744&0.564&0.640&0.563&0.534&0.427&0.455\\\hline
DEF&\color{red}{\textbf{0.817}}&0.748&\color{blue}{\underline{0.768}}&0.754&0.751&0.521&0.586&0.512&0.487&0.455&0.409\\\hline
MB&\color{blue}{\underline{0.663}}&\color{red}{\textbf{0.783}}&0.578&0.654&0.621&0.551&0.339&0.518&0.375&0.357&0.222\\\hline
FM&\color{blue}{\underline{0.643}}&\color{red}{\textbf{0.745}}&0.575&0.612&0.581&0.604&0.333&0.551&0.353&0.396&0.220\\\hline
IPR&\color{blue}{\underline{0.783}}&\color{red}{\textbf{0.795}}&0.706&0.739&0.731&0.617&0.597&0.584&0.579&0.453&0.457\\\hline
OPR&\color{red}{\textbf{0.831}}&\color{blue}{\underline{0.807}}&0.741&0.741&0.724&0.597&0.618&0.596&0.604&0.466&0.464\\\hline
OV&\color{red}{\textbf{0.654}}&\color{blue}{\underline{0.641}}&0.495&0.584&0.555&0.539&0.429&0.576&0.455&0.393&0.307\\\hline
BC&\color{blue}{\underline{0.789}}&\color{red}{\textbf{0.840}}&0.761&0.698&0.725&0.585&0.578&0.428&0.578&0.456&0.421\\\hline
LR&0.369&0.478&\color{blue}{\underline{0.539}}&0.516&0.379&\color{red}{\textbf{0.545}}&0.305&0.349&0.187&0.171&0.278\\\hline
\hline
\end{tabular}
\label{tab:subsets}
\end{table*}

\subsection{Comparison of results on ALOV300++}\label{sec::alov}
To further validate the robustness of the BIT, we conduct the second evaluation on a larger dataset \cite{alov}, namely the Amsterdam Library of Ordinary Videos (ALOV300++), which was recently developed by Smeulders et al.. It consists of 14 challenge subsets, with 315 sequences in total, and focuses on systematically and experimentally evaluating the robustness of trackers in a large variety of situations including light changes, low contrast, occlusion, etc. In \cite{alov}, survival curves based on F-score were proposed to evaluate each tracker’s robustness and demonstrate its effectiveness. To obtain the survival curve of a tracker, a F-score for each video is computed $F = 2\times(precision\times recall)/(precision + recall)$, where $precision = ntp/(ntp + nfp)$, $recall = ntp/(ntp +nfn)$, and $ntp$, $nfp$, $nfn$ respectively denote the number of true positives, false positives and false negatives in a video. A reasonable choice for the overlap of target and object is the PASCAL criterion \cite{pascal}: $\left|{T^i\bigcap GT^i}\right|/\left|{T^i\bigcup GT^i}\right|\geq 0.5$, where $T^i$ denotes the tracked bounding box in frame $i$, and $GT^i$ denotes the ground truth bounding box in frame $i$. A survival curve shows the performance of a tracker on all videos in the dataset. The videos are sorted according to the F-score, and the graph gives a bird’s eye view of the cumulative rendition of the quality of the tracker on the whole dataset.

\begin{figure}[!t]
\centering
\includegraphics[width=1.0\linewidth]{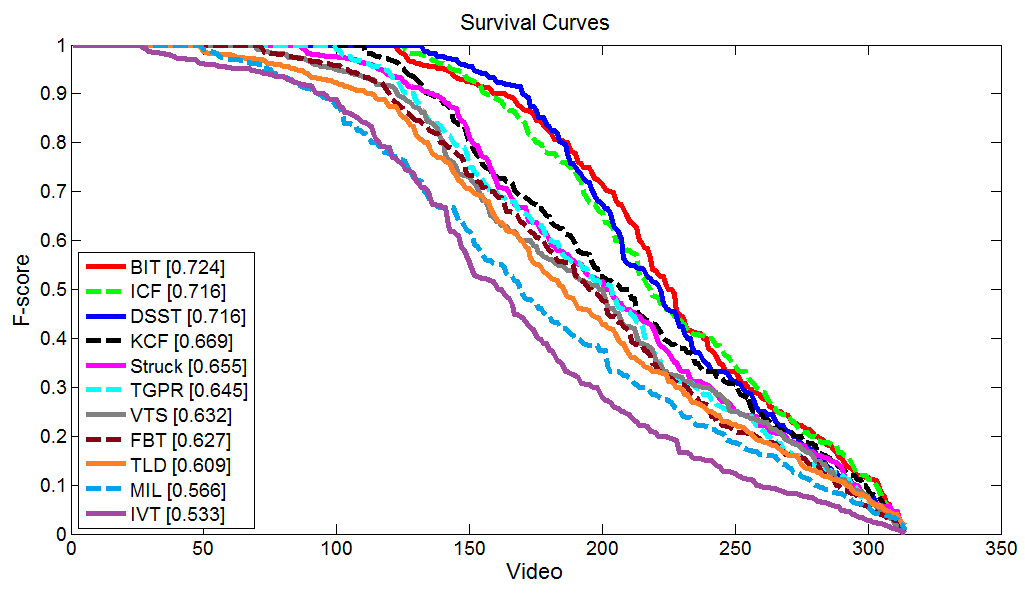}
\caption{The survival curves for the top ten trackers on the ALOV300++}
\label{fig:sc}
\end{figure}

To evaluate the BIT on the ALOV300++ dataset \cite{alov}, we ran the BIT on all 315 sequences using the ground truth of the first frame as the initialization and the same parameters as the previous evaluation. Because the F-score is also sensitive to bounding box size, DSST \cite{ddst} is only used to estimate bounding box size same as the success plot in Section \ref{sec:CVPR2013}. We compare our tracker with the top four popular trackers (Struck \cite{struck}, VTS \cite{vts}, FBT \cite{fbt}, TLD \cite{tld}) which were evaluated in \cite{alov}. In addition, we also run ICF \cite{icf}, DSST\cite{ddst}, KCF \cite{kcf} and TGPR \cite{tgpr} on ALOV300++, which are recognized as the state-of-art trackers in the previous evaluation, and two classical trackers (IVT \cite{ivt} and MIL \cite{mil}) are used as comparison baselines. The survival curves of the ten trackers and the average F-scores over all sequences are shown in Figure \ref{fig:sc}, which demonstrates that the BIT achieves the best overall performance compared to the 10 trackers in this comparison. From the figure, we can see that the proposed tracker outperforms ICF \cite{icf} by 0.008 in mean F-score and the survival rate ($F>0.8$) of the BIT is 59.36\% compared to the second best tracker 55.87\% - a difference of 3.49\%.

\subsection{Tracking model analysis}
\begin{figure}[!t]
\centering
\includegraphics[width=0.675\linewidth]{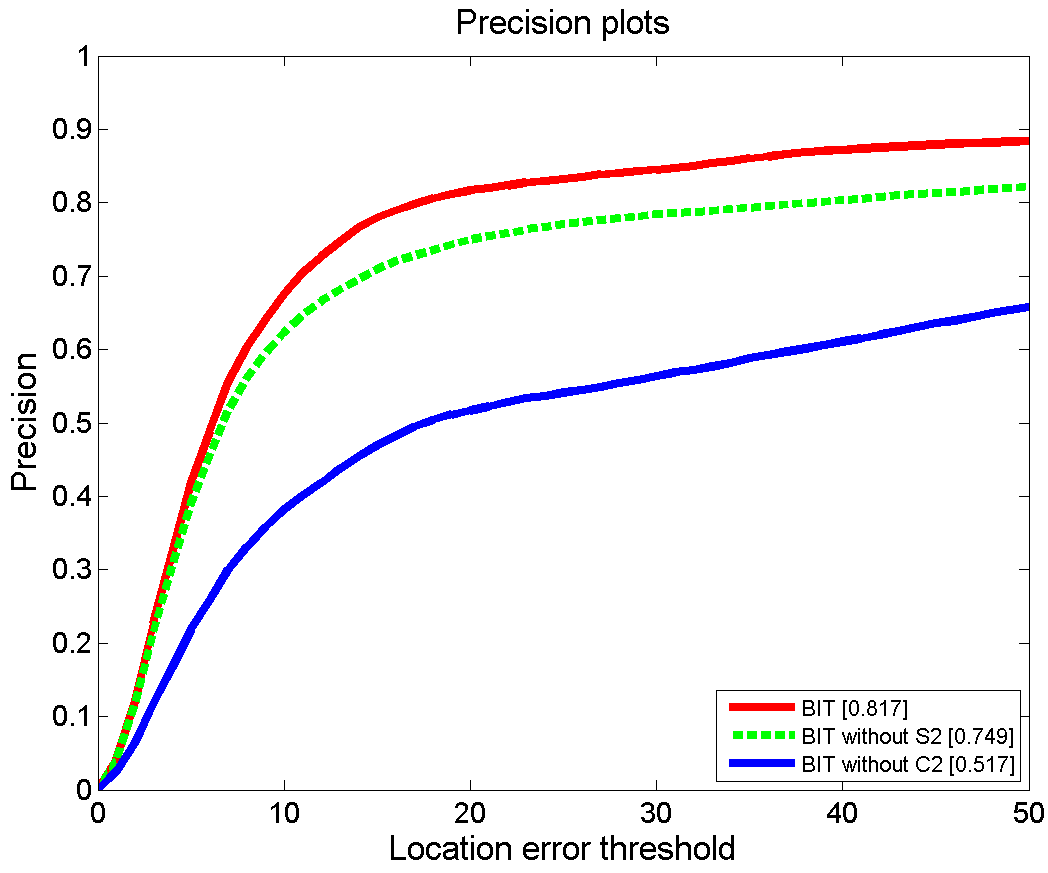}
\caption{Generative and discriminative model analysis on the TB2013}
\label{fig:modelanalysis}
\end{figure}

According to the introduction, tracking model can be generative or discriminative. For generative models, tracker searches for the most similar region to the target object within a neighborhood. For discriminative models, tracker is a classifier to distinguish the target object from the background. However, generative models ignore negative samples in the background and discriminative models pay insufficient attention to the eigen basis of the tracking target. In this paper, BIT proposed for advanced learning combines the generative and discriminative model: the response of view-tuned learning (S2 units) is a generative model and a fully-connected network classifier simulates for task-turned learning (C2 units) as a discriminative model. In order to prove the effect of hybrid-model, we compare the tracking scores between BIT without C2 units, BIT without S2 units and intact BIT on the TB2013 showed in Fig. \ref{fig:modelanalysis}. For the generative model only (without C2 units), the object location $\left(\hat{x},\hat{y}\right)$ is determined by maximizing the S2 response map as $\left( {\hat x,\hat y} \right) = \arg {\min _{\left( {x,y} \right)}}S2\left( {x,y} \right)$. For the discriminative model only (without S2 units), the convolutional neural network in C2 units receives the activity from C1 units directly as $C2\left( {x,y} \right) = {1/K}\sum_{k} {{\mathcal{F}^{ - 1}}\left[ {\mathcal{F}\left[ {W\left( {x,y,k} \right)} \right] \odot \mathcal{F}\left[ {C1\left( {x,y,k} \right)} \right]} \right]}$. Clearly, the hybrid-model (81.7\%) achieved excellent performances in comparison to single-model (74.9\% and 51.7\%). In addition, the performance gap between the discriminative model and the generative model in the literature is 23.2\% for the tracking precision measure. Because background information is critical for effective tracking, which can be exploited by using advanced learning techniques to encode the background information in the discriminative model implicitly or serving as the tracking context explicitly.

\subsection{Tracking speed analysis}
The experiments were run on a PC with Intel i7 3770 CPU (3.4GHz), and we report the average speed (in fps) of the proposed BIT method in Table \ref{tab:fps} and compare it with the state-of-the-art visual trackers referred to in Sec. \ref{sec:CVPR2013}. In this paper, the average speed score is defined as the average fps over all the sequences, which objectively evaluates sequences where the initialization process usually dominates the computational burden. According to the standard frame rate of films, we consider that a processing speed of more than 24 fps is equivalent to real-time processing.

\begin{figure}
\centering
\includegraphics[width=0.8\linewidth]{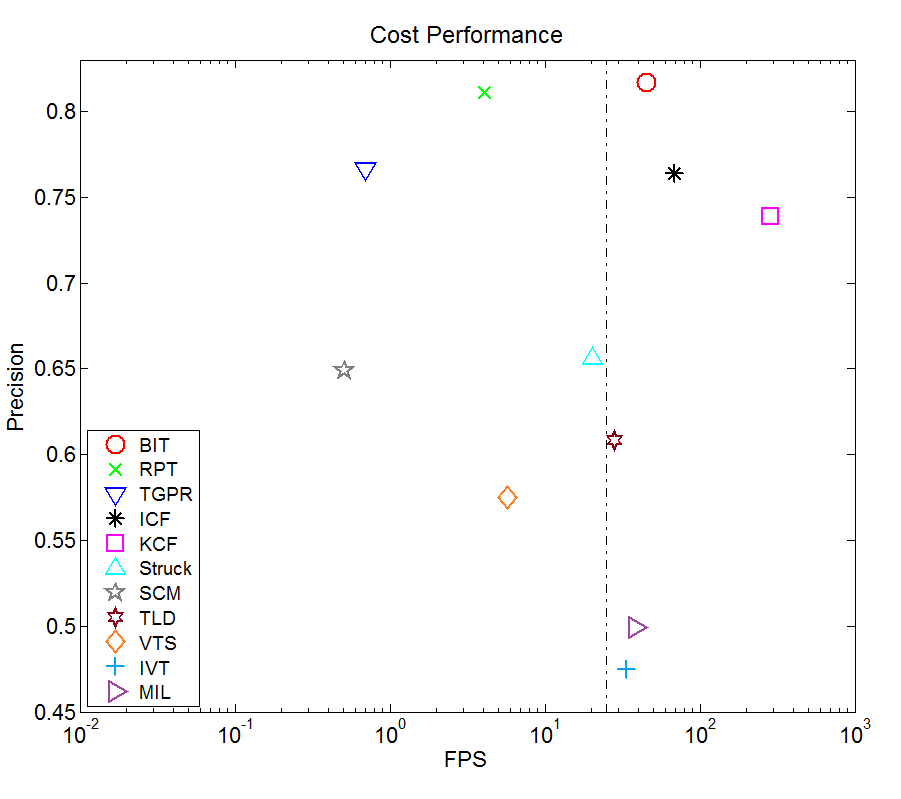}
\caption{Cost performance on the TB2013. The dashed line is the boundary between real-time and non-real-time.}
\label{fig:fps}
\end{figure}

Table \ref{tab:fps} and Fig.\ref{fig:fps} show the tracking speed and the tracking scores on the TB2013 \cite{cvpr2013}. According to the table, our method tracks the object at an average speed of 45fps, which is significantly faster than the second best tracker RPT (4.1 fps). Furthermore, the speed of our proposed BIT is close to twice that of the real-time criterion, which leaves a substantial amount of time in which to increase modified strategies to further improve tracking performance.
\begin{table}
\center
\caption{The tracking speed and scores on the TB2013 (M: Matlab, C:C/C++, MC: Mixture of Matlab and C/C++, suffix E: executable binary code).}
\footnotesize
\begin{tabular}{c|c|c|c|c}
\hline
\hline
& Tracker & Speed (fps)& Precision (\%)&Code\\\hline
\multirow{6}*{Real-time}&BIT&44.9&\textbf{81.7}&MC\\
&ICF\cite{icf}&68.8&76.4&MC\\
&KCF\cite{kcf}&\textbf{284.4}&73.9&MC\\
&TLD\cite{tld}&28.1&60.8&MC\\
&MIL\cite{mil}&38.1&49.9&C\\
&IVT\cite{ivt}&33.4&47.5&MC\\
\hline
\multirow{5}*{Non real-time}&RPT\cite{rpt}&4.1&81.1&MC\\
&TGPR\cite{tgpr}&0.7&76.6&C\\
&Struck\cite{struck}&20.2&65.6&C\\
&SCM\cite{scm}&0.5&64.9&MC\\
&VTS\cite{vts}&5.7&57.5&MC-E\\
\hline\hline
\end{tabular}
\label{tab:fps}
\end{table}

\section{Conclusion}\label{sec:conclusion}
In this paper, we successfully applied a bio-inspired model to real-time visual tracking. To the best of our knowledge, this is the first time this has been achieved. Inspired by bioresearch, the proposed novel bio-inspired tracker models the ventral stream of the primate visual cortex, extracting low-level bio-inspired features in S1 and C1 units, and simulating high-level learning mechanisms in S2 and C2 units. Furthermore, the complicated bio-inspired tracker operates in real-time since fast Gabor approximation (FGA) and fast Fourier transform (FFT) are used for online learning and detection. Numerous experiments with state-of-the-art algorithms on challenging sequences demonstrate that the BIT achieves favorable results in terms of robustness and speed.

The human visual system is a complex neural network which is multi-layered and multi-synaptic, and the BIT pays little attention to a number of visual cognitive mechanisms. A well-known multi-store memory model (which includes sensory memory, short-term memory, and long-term memory) is proposed by Atkinson and Shiffrin \cite{asmm}, which provides some ideas for tracking-by-detection to improve out-of-view and long-term challenges. Moreover, a multi-stored prototype \cite{hmax} is used for object recognition in S2 units, which has proved to be effective for local occlusion and partial deformation. In addition, C2 units using a single layer convolutional network cannot perfectly simulate neuron connections in PFC, which provides a good starting platform for further research into multi-layer neural networks.

\section*{Acknowledgment}
This work is supported by the National Natural Science Foundation of China (\#61171142, \#61401163), the Science and technology Planning Project of Guangdong Province of China (\#2011A010801005, \#2012B061700102, \#2014B010111003, \#2014B010111006) and Australian Research Council Projects (FT-130101457 and DP-140102164).

\ifCLASSOPTIONcaptionsoff
  \newpage
\fi

\bibliographystyle{IEEEtran}
\bibliography{bit}

\begin{thebibliography}{10}
\providecommand{\url}[1]{#1}
\csname url@samestyle\endcsname
\providecommand{\newblock}{\relax}
\providecommand{\bibinfo}[2]{#2}
\providecommand{\BIBentrySTDinterwordspacing}{\spaceskip=0pt\relax}
\providecommand{\BIBentryALTinterwordstretchfactor}{4}
\providecommand{\BIBentryALTinterwordspacing}{\spaceskip=\fontdimen2\font plus
\BIBentryALTinterwordstretchfactor\fontdimen3\font minus
  \fontdimen4\font\relax}
\providecommand{\BIBforeignlanguage}[2]{{%
\expandafter\ifx\csname l@#1\endcsname\relax
\typeout{** WARNING: IEEEtran.bst: No hyphenation pattern has been}%
\typeout{** loaded for the language `#1'. Using the pattern for}%
\typeout{** the default language instead.}%
\else
\language=\csname l@#1\endcsname
\fi
#2}}
\providecommand{\BIBdecl}{\relax}
\BIBdecl

\bibitem{review}
K.~Cannons, ``A review of visual tracking,'' \emph{Dept. Comput. Sci. Eng.,
  York Univ., Toronto, Canada, Tech. Rep. CSE-2008-07}, 2008.

\bibitem{kms}
D.~Comaniciu, V.~Ramesh, and P.~Meer, ``Kernel-based object tracking,''
  \emph{Pattern Analysis and Machine Intelligence}, vol.~25, no.~5, pp.
  564--577, 2003.

\bibitem{kcf}
J.~F. Henriques, R.~Caseiro, P.~Martins, and J.~Batista, ``High-speed tracking
  with kernelized correlation filters,'' \emph{Pattern Analysis and Machine
  Intelligence}, vol.~37, no.~3, pp. 583--596, 2015.

\bibitem{dft}
L.~Sevilla-Lara and E.~Learned-Miller, ``Distribution fields for tracking,'' in
  \emph{Computer Vision and Pattern Recognition}.\hskip 1em plus 0.5em minus
  0.4em\relax IEEE, 2012, pp. 1910--1917.

\bibitem{cxt}
T.~B. Dinh, N.~Vo, and G.~Medioni, ``Context tracker: Exploring supporters and
  distracters in unconstrained environments,'' in \emph{Computer Vision and
  Pattern Recognition}.\hskip 1em plus 0.5em minus 0.4em\relax IEEE, 2011, pp.
  1177--1184.

\bibitem{sift}
H.~Zhou, Y.~Yuan, and C.~Shi, ``Object tracking using sift features and mean
  shift,'' \emph{Computer vision and image understanding}, vol. 113, no.~3, pp.
  345--352, 2009.

\bibitem{ct}
K.~Zhang, L.~Zhang, and M.-H. Yang, ``Real-time compressive tracking,'' in
  \emph{Computer Vision--ECCV}.\hskip 1em plus 0.5em minus 0.4em\relax
  Springer, 2012, pp. 864--877.

\bibitem{selftaught}
R.~Raina, A.~Battle, H.~Lee, B.~Packer, and A.~Y. Ng, ``Self-taught learning:
  transfer learning from unlabeled data,'' in \emph{Proceedings of the 24th
  international conference on Machine learning}.\hskip 1em plus 0.5em minus
  0.4em\relax ACM, 2007, pp. 759--766.

\bibitem{ivt}
D.~A. Ross, J.~Lim, R.-S. Lin, and M.-H. Yang, ``Incremental learning for
  robust visual tracking,'' \emph{International Journal of Computer Vision},
  vol.~77, no. 1-3, pp. 125--141, 2008.

\bibitem{lda}
R.-S. Lin, M.-H. Yang, and S.~E. Levinson, ``Object tracking using incremental
  fisher discriminant analysis,'' in \emph{Pattern Recognition, 2004. ICPR
  2004. Proceedings of the 17th International Conference on}, vol.~2.\hskip 1em
  plus 0.5em minus 0.4em\relax IEEE, 2004, pp. 757--760.

\bibitem{llc}
G.~Wang, X.~Qin, F.~Zhong, Y.~Liu, Q.~Peng, and M.-H. Yang, ``Visual tracking
  via sparse and local linear coding,'' \emph{Image Processing, IEEE
  Transactions on}, vol.~24, no.~11, pp. 3796--3809, 2015.

\bibitem{scm}
W.~Zhong, H.~Lu, and M.-H. Yang, ``Robust object tracking via sparsity-based
  collaborative model,'' in \emph{Computer Vision and Pattern
  Recognition}.\hskip 1em plus 0.5em minus 0.4em\relax IEEE, 2012, pp.
  1838--1845.

\bibitem{mtmvt}
Z.~Hong, X.~Mei, D.~Prokhorov, and D.~Tao, ``Tracking via robust multi-task
  multi-view joint sparse representation,'' in \emph{Computer Vision (ICCV),
  2013 IEEE International Conference on}.\hskip 1em plus 0.5em minus
  0.4em\relax IEEE, 2013, pp. 649--656.

\bibitem{mvt}
X.~Mei, Z.~Hong, D.~Prokhorov, and D.~Tao, ``Robust multitask multiview
  tracking in videos,'' \emph{IEEE Transactions on Neural Networks and Learning
  Systems}, vol.~26, no.~11, pp. 2874--2890, 2015.

\bibitem{sfa}
S.~Liwicki, S.~Zafeiriou, and M.~Pantic, ``Online kernel slow feature analysis
  for temporal video segmentation and tracking,'' \emph{Image Processing IEEE
  Transactions on}, vol.~24, p.~1, 2015.

\bibitem{simultaneous}
N.~M. Nayak, Y.~Zhu, and A.~K. Roy~Chowdhury, ``Hierarchical graphical models
  for simultaneous tracking and recognition in wide-area scenes,'' \emph{Image
  Processing, IEEE Transactions on}, vol.~24, no.~7, pp. 2025--2036, 2015.

\bibitem{ht}
M.~Godec, P.~M. Roth, and H.~Bischof, ``Hough-based tracking of non-rigid
  objects,'' \emph{Computer Vision and Image Understanding}, vol. 117, no.~10,
  pp. 1245--1256, 2013.

\bibitem{asla}
X.~Jia, H.~Lu, and M.-H. Yang, ``Visual tracking via adaptive structural local
  sparse appearance model,'' in \emph{Computer Vision and Pattern
  Recognition}.\hskip 1em plus 0.5em minus 0.4em\relax IEEE, 2012, pp.
  1822--1829.

\bibitem{dsat}
J.~Fan, Y.~Wu, and S.~Dai, ``Discriminative spatial attention for robust
  tracking,'' in \emph{Computer Vision--ECCV}.\hskip 1em plus 0.5em minus
  0.4em\relax Springer, 2010, pp. 480--493.

\bibitem{discriminative}
Y.~Wu, M.~Pei, M.~Yang, J.~Yuan, and Y.~Jia, ``Robust discriminative tracking
  via landmark-based label propagation,'' \emph{Image Processing, IEEE
  Transactions on}, vol.~24, no.~5, pp. 1510--1523, 2015.

\bibitem{nonrigid}
X.~Sun, H.~Yao, S.~Zhang, and D.~Li, ``Non-rigid object contour tracking via a
  novel supervised level set model,'' \emph{Image Processing, IEEE Transactions
  on}, vol.~24, no.~11, pp. 3386--3399, 2015.

\bibitem{msvm}
S.~Zhang, X.~Yu, Y.~Sui, S.~Zhao, and L.~Zhang, ``Object tracking with
  multi-view support vector machines,'' \emph{Multimedia, IEEE Transactions
  on}, vol.~17, no.~3, pp. 265--278, 2015.

\bibitem{struck}
S.~Hare, A.~Saffari, and P.~H. Torr, ``Struck: Structured output tracking with
  kernels,'' in \emph{International Conference on Computer Vision}.\hskip 1em
  plus 0.5em minus 0.4em\relax IEEE, 2011, pp. 263--270.

\bibitem{oab}
H.~Grabner, M.~Grabner, and H.~Bischof, ``Real-time tracking via on-line
  boosting.'' in \emph{BMVC}, vol.~1, no.~5, 2006, p.~6.

\bibitem{mil}
B.~Babenko, M.-H. Yang, and S.~Belongie, ``Visual tracking with online multiple
  instance learning,'' in \emph{Computer Vision and Pattern Recognition}.\hskip
  1em plus 0.5em minus 0.4em\relax IEEE, 2009, pp. 983--990.

\bibitem{dml}
Z.~Hong, X.~Mei, and D.~Tao, ``Dual-force metric learning for robust
  distracter-resistant tracker,'' in \emph{Computer Vision--ECCV 2012}.\hskip
  1em plus 0.5em minus 0.4em\relax Springer, 2012, pp. 513--527.

\bibitem{cgdt}
Q.~Yu, T.~B. Dinh, and G.~Medioni, ``Online tracking and reacquisition using
  co-trained generative and discriminative trackers,'' in \emph{Computer
  Vision--ECCV}.\hskip 1em plus 0.5em minus 0.4em\relax Springer, 2008, pp.
  678--691.

\bibitem{objectrecognition}
Y.~Huang, K.~Huang, D.~Tao, T.~Tan, and X.~Li, ``Enhanced biologically inspired
  model for object recognition,'' \emph{Systems, Man, and Cybernetics, Part B:
  Cybernetics, IEEE Transactions on}, vol.~41, no.~6, pp. 1668--1680, 2011.

\bibitem{faceidentification}
B.~Ma, Y.~Su, and F.~Jurie, ``Covariance descriptor based on bio-inspired
  features for person re-identification and face verification,'' \emph{Image
  and Vision Computing}, vol.~32, no.~6, pp. 379--390, 2014.

\bibitem{sceneclassification}
D.~Song and D.~Tao, ``Biologically inspired feature manifold for scene
  classification,'' \emph{Image Processing, IEEE Transactions on}, vol.~19,
  no.~1, pp. 174--184, 2010.

\bibitem{cvpr2013}
Y.~Wu, J.~Lim, and M.-H. Yang, ``Online object tracking: A benchmark,'' in
  \emph{Computer Vision and Pattern Recognition}.\hskip 1em plus 0.5em minus
  0.4em\relax IEEE, 2013, pp. 2411--2418.

\bibitem{alov}
A.~W. Smeulders, D.~M. Chu, R.~Cucchiara, S.~Calderara, A.~Dehghan, and
  M.~Shah, ``Visual tracking: an experimental survey,'' \emph{Pattern Analysis
  and Machine Intelligence}, vol.~36, no.~7, pp. 1442--1468, 2014.

\bibitem{hierarchicalmodel}
M.~Riesenhuber and T.~Poggio, ``Hierarchical models of object recognition in
  cortex,'' \emph{Nature neuroscience}, vol.~2, no.~11, pp. 1019--1025, 1999.

\bibitem{hmax}
T.~Serre and M.~Riesenhuber, ``Realistic modeling of simple and complex cell
  tuning in the hmax model, and implications for invariant object recognition
  in cortex,'' DTIC Document, Tech. Rep., 2004.

\bibitem{imagenetdcnn}
A.~Krizhevsky, I.~Sutskever, and G.~E. Hinton, ``Imagenet classification with
  deep convolutional neural networks,'' in \emph{Advances in neural information
  processing systems}, 2012, pp. 1097--1105.

\bibitem{rfh}
R.~Girshick, J.~Donahue, T.~Darrell, and J.~Malik, ``Rich feature hierarchies
  for accurate object detection and semantic segmentation,'' in \emph{Computer
  Vision and Pattern Recognition}.\hskip 1em plus 0.5em minus 0.4em\relax IEEE,
  2014, pp. 580--587.

\bibitem{cnnhumantracking}
J.~Fan, W.~Xu, Y.~Wu, and Y.~Gong, ``Human tracking using convolutional neural
  networks,'' \emph{Neural Networks, IEEE Transactions on}, vol.~21, no.~10,
  pp. 1610--1623, 2010.

\bibitem{dlt}
N.~Wang and D.-Y. Yeung, ``Learning a deep compact image representation for
  visual tracking,'' in \emph{Advances in Neural Information Processing
  Systems}, 2013, pp. 809--817.

\bibitem{80mti}
A.~Torralba, R.~Fergus, and W.~T. Freeman, ``80 million tiny images: A large
  data set for nonparametric object and scene recognition,'' \emph{Pattern
  Analysis and Machine Intelligence}, vol.~30, no.~11, pp. 1958--1970, 2008.

\bibitem{dst}
V.~Mahadevan and N.~Vasconcelos, ``Biologically inspired object tracking using
  center-surround saliency mechanisms,'' \emph{Pattern Analysis and Machine
  Intelligence, IEEE Transactions on}, vol.~35, no.~3, pp. 541--554, 2013.

\bibitem{pso}
Y.~Zheng and Y.~Meng, ``Swarm intelligence based dynamic object tracking,'' in
  \emph{Evolutionary Computation, 2008. CEC 2008.(IEEE World Congress on
  Computational Intelligence). IEEE Congress on}.\hskip 1em plus 0.5em minus
  0.4em\relax IEEE, 2008, pp. 405--412.

\bibitem{sbif}
M.~Li, Z.~Zhang, K.~Huang, and T.~Tan, ``Robust visual tracking based on
  simplified biologically inspired features,'' in \emph{ICIP}.\hskip 1em plus
  0.5em minus 0.4em\relax IEEE, 2009, pp. 4113--4116.

\bibitem{cnnt}
K.~Zhang, Q.~Liu, Y.~Wu, and M.-H. Yang, ``Robust tracking via convolutional
  networks without learning,'' \emph{arXiv preprint arXiv:1501.04505}, 2015.

\bibitem{2dvcf}
J.~G. Daugman, ``Uncertainty relation for resolution in space, spatial
  frequency, and orientation optimized by two-dimensional visual cortical
  filters,'' \emph{JOSA A}, vol.~2, no.~7, pp. 1160--1169, 1985.

\bibitem{cortex-like}
T.~Serre, L.~Wolf, S.~Bileschi, M.~Riesenhuber, and T.~Poggio, ``Robust object
  recognition with cortex-like mechanisms,'' \emph{Pattern Analysis and Machine
  Intelligence}, vol.~29, no.~3, pp. 411--426, 2007.

\bibitem{itti}
C.~Siagian and L.~Itti, ``Rapid biologically-inspired scene classification
  using features shared with visual attention,'' \emph{Pattern Analysis and
  Machine Intelligence}, vol.~29, no.~2, pp. 300--312, 2007.

\bibitem{saliency}
X.~Cao, Z.~Tao, B.~Zhang, H.~Fu, and W.~Feng, ``Self-adaptively weighted
  co-saliency detection via rank constraint,'' \emph{Image Processing, IEEE
  Transactions on}, vol.~23, no.~9, pp. 4175--4186, 2014.

\bibitem{doubleopponent}
L.~Itti, C.~Koch, and E.~Niebur, ``A model of saliency-based visual attention
  for rapid scene analysis,'' \emph{IEEE Transactions on Pattern Analysis \&
  Machine Intelligence}, no.~11, pp. 1254--1259, 1998.

\bibitem{cn}
J.~Van De~Weijer, C.~Schmid, J.~Verbeek, and D.~Larlus, ``Learning color names
  for real-world applications,'' \emph{Image Processing, IEEE Transactions on},
  vol.~18, no.~7, pp. 1512--1523, 2009.

\bibitem{catsfields}
D.~H. Hubel and T.~N. Wiesel, ``Receptive fields, binocular interaction and
  functional architecture in the cat's visual cortex,'' \emph{The Journal of
  physiology}, vol. 160, no.~1, p. 106, 1962.

\bibitem{maxpool}
I.~Lampl, D.~Ferster, T.~Poggio, and M.~Riesenhuber, ``Intracellular
  measurements of spatial integration and the max operation in complex cells of
  the cat primary visual cortex,'' \emph{Journal of neurophysiology}, vol.~92,
  no.~5, pp. 2704--2713, 2004.

\bibitem{age}
G.~Guo, G.~Mu, Y.~Fu, and T.~S. Huang, ``Human age estimation using
  bio-inspired features,'' in \emph{Computer Vision and Pattern
  Recognition}.\hskip 1em plus 0.5em minus 0.4em\relax IEEE, 2009, pp.
  112--119.

\bibitem{hog}
N.~Dalal and B.~Triggs, ``Histograms of oriented gradients for human
  detection,'' in \emph{Computer Vision and Pattern Recognition}, vol.~1.\hskip
  1em plus 0.5em minus 0.4em\relax IEEE, 2005, pp. 886--893.

\bibitem{bio}
T.~Serre, M.~Kouh, C.~Cadieu, U.~Knoblich, G.~Kreiman, and T.~Poggio, ``A
  theory of object recognition: computations and circuits in the feedforward
  path of the ventral stream in primate visual cortex,'' DTIC Document, Tech.
  Rep., 2005.

\bibitem{tld}
Z.~Kalal, J.~Matas, and K.~Mikolajczyk, ``P-n learning: Bootstrapping binary
  classifiers by structural constraints,'' in \emph{Computer Vision and Pattern
  Recognition}.\hskip 1em plus 0.5em minus 0.4em\relax IEEE, 2010, pp. 49--56.

\bibitem{fft}
H.~J. Nussbaumer, \emph{Fast Fourier transform and convolution
  algorithms}.\hskip 1em plus 0.5em minus 0.4em\relax Springer Science \&
  Business Media, 2012, vol.~2.

\bibitem{signals}
S.~Haykin and B.~Van~Veen, \emph{Signals and systems}.\hskip 1em plus 0.5em
  minus 0.4em\relax John Wiley \& Sons, 2007.

\bibitem{rpt}
Y.~Li, J.~Zhu, and S.~C. Hoi, ``Reliable patch trackers: Robust visual tracking
  by exploiting reliable patches,'' in \emph{Computer Vision and Pattern
  Recognition}, June 2015.

\bibitem{tgpr}
J.~Gao, H.~Ling, W.~Hu, and J.~Xing, ``Transfer learning based visual tracking
  with gaussian processes regression,'' in \emph{Computer Vision--ECCV}.\hskip
  1em plus 0.5em minus 0.4em\relax Springer, 2014, pp. 188--203.

\bibitem{icf}
Z.~Hong, Z.~Chen, C.~Wang, X.~Mei, D.~Prokhorov, and D.~Tao, ``Multi-store
  tracker (muster): A cognitive psychology inspired approach to object
  tracking,'' in \emph{Computer Vision and Pattern Recognition}, June 2015.

\bibitem{vts}
J.~Kwon and K.~M. Lee, ``Tracking by sampling trackers,'' in \emph{ICCV}.\hskip
  1em plus 0.5em minus 0.4em\relax IEEE, 2011, pp. 1195--1202.

\bibitem{ddst}
M.~Danelljan, G.~H{\"a}ger, F.~Khan, and M.~Felsberg, ``Accurate scale
  estimation for robust visual tracking,'' in \emph{British Machine Vision
  Conference, Nottingham, September 1-5, 2014}.\hskip 1em plus 0.5em minus
  0.4em\relax BMVA Press, 2014.

\bibitem{pascal}
M.~Everingham, L.~Van~Gool, C.~K. Williams, J.~Winn, and A.~Zisserman, ``The
  pascal visual object classes (voc) challenge,'' \emph{International journal
  of computer vision}, vol.~88, no.~2, pp. 303--338, 2010.

\bibitem{fbt}
H.~T. Nguyen and A.~W. Smeulders, ``Robust tracking using foreground-background
  texture discrimination,'' \emph{International Journal of Computer Vision},
  vol.~69, no.~3, pp. 277--293, 2006.

\bibitem{asmm}
R.~C. Atkinson and R.~M. Shiffrin, ``Human memory: A proposed system and its
  control processes,'' \emph{The psychology of learning and motivation},
  vol.~2, pp. 89--195, 1968.

\end{thebibliography}

\begin{IEEEbiography}[{\includegraphics[width=1in,height=1.25in,clip,keepaspectratio]{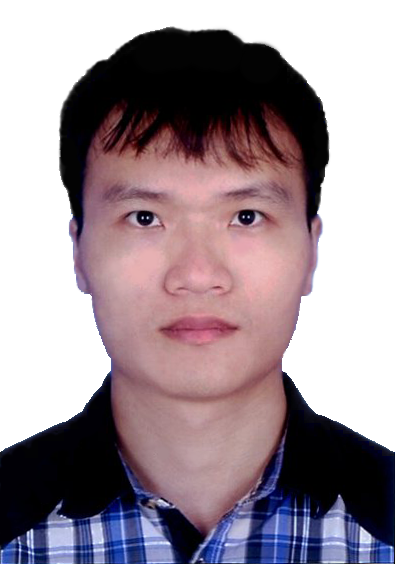}}]{Bolun Cai}
 received his B.S. degree from South China University of Technology, China, in 2013. Currently, he is a master student in School of Electronic and Information Engineering, South China University of Technology, China. His research interests include image processing, object tracking/recognition and machine learning.
\end{IEEEbiography}

\begin{IEEEbiography}[{\includegraphics[width=1in,height=1.25in,clip,keepaspectratio]{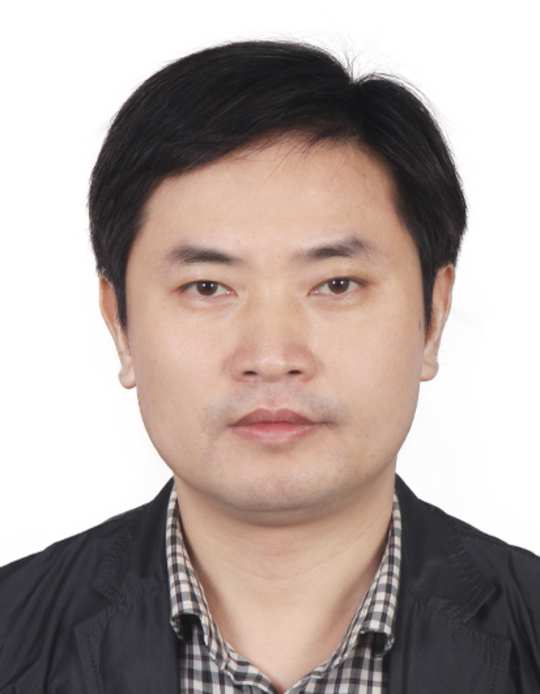}}]{Xiangmin Xu}
 received the Ph.D. degree from the South China University of Technology, Guangzhou, China. He is currently a Full Professor with the School of Electronic and Information Engineering, South China University of Technology, Guangzhou, China. His current research interests include image/video processing, human-computer interaction, computer vision, and machine learning.
\end{IEEEbiography}

\begin{IEEEbiography}[{\includegraphics[width=1in,height=1.25in,clip,keepaspectratio]{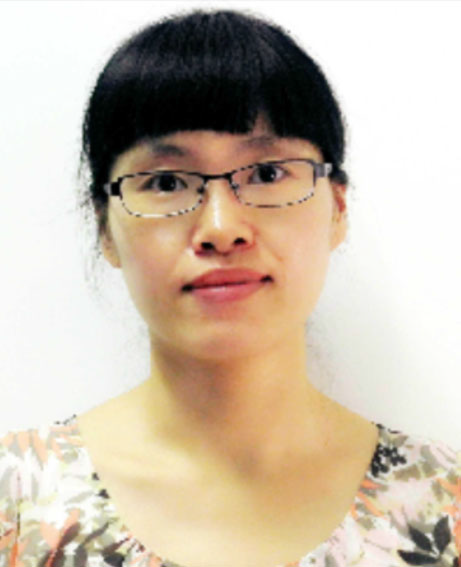}}]{Xiaofen Xing}
 received her B.S. degree, M.S. degree and Ph.D. degree from South China University of Technology, in 2001, 2004 and 2013, respectively. From 2007 till now, she is a lecturer in School of Electronic and Information Engineering, South China University of Technology, China. Her main research interests include image/video processing, human-computer interaction and video surveillance.
\end{IEEEbiography}

\begin{IEEEbiography}[{\includegraphics[width=1in,height=1.25in,clip,keepaspectratio]{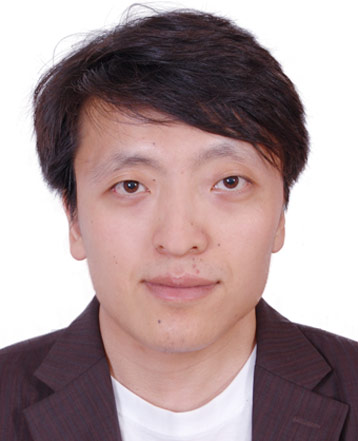}}]{Kui Jia}
 received the B.Eng. degree in marine engineering from Northwestern Polytechnic University, China, in 2001, the M.Eng. degree in electrical and computer engineering from National University of Singapore in 2003, and the Ph.D. degree in computer science from Queen Mary, University of London, London, U.K., in 2007. He is currently a Visiting Assistant Professor at University of Macau, Macau SAR, China. His research interests are in computer vision, machine learning, and image processing.
\end{IEEEbiography}

\begin{IEEEbiography}[{\includegraphics[width=1in,height=1.25in,clip,keepaspectratio]{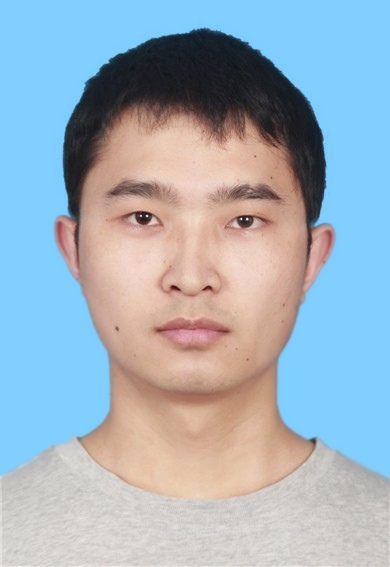}}]{Miao Jie}
received the B.S. degree from the School of Electronic and Information Engineering, South China University of Technology, Guangzhou, China, in 2010, where he is currently pursuing the Ph.D. degree with the School of Electronic and Information Engineering. His research interests include spatial-temporal features, unsupervised feature learning, and action recognition.
\end{IEEEbiography}

\begin{IEEEbiography}[{\includegraphics[width=1in,height=1.25in,clip,keepaspectratio]{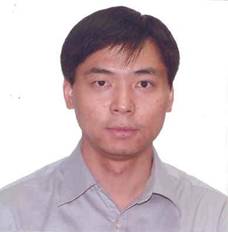}}]{Dacheng Tao}
(F'15) is Professor of Computer Science with the Centre for Quantum Computation \& Intelligent Systems, and the Faculty of Engineering and Information Technology in the University of Technology Sydney. He mainly applies statistics and mathematics to data analytics problems and his research interests spread across computer vision, data science, image processing, machine learning, and video surveillance. His research results have expounded in one monograph and 200+ publications at prestigious journals and prominent conferences, such as IEEE T-PAMI, T-NNLS, T-IP, JMLR, IJCV, NIPS, ICML, CVPR, ICCV, ECCV, AISTATS, ICDM; and ACM SIGKDD, with several best paper awards, such as the best theory/algorithm paper runner up award in IEEE ICDM'07, the best student paper award in IEEE ICDM'13, and the 2014 ICDM 10-year highest-impact paper award. He received the 2015 Australian Scopus-Eureka Prize, the 2015 ACS Gold Disruptor Award and the 2015 UTS Vice-Chancellor's Medal for Exceptional Research. He is a Fellow of the IEEE, OSA, IAPR and SPIE.
\end{IEEEbiography}



\end{document}